\pdfoutput=1

\documentclass[11pt]{article}

\usepackage[final]{acl}

\usepackage{times}
\usepackage{latexsym}
\usepackage{hyperref}
\usepackage{url}
\usepackage{inconsolata}
\usepackage{multirow}
\usepackage{graphicx}
\usepackage{color}
\usepackage{xcolor}
\usepackage{amsmath}
\usepackage{enumitem}
\usepackage{mdframed}
\usepackage{bm}
\usepackage{arydshln}
\usepackage{tcolorbox}
\usepackage{algorithm}
\usepackage{algorithmicx}
\usepackage[noend]{algpseudocode}
\usepackage{subcaption}
\usepackage[export]{adjustbox}
\usepackage{rotating}
\usepackage{booktabs}
\usepackage{marvosym}
\usepackage{footmisc}
\usepackage{amssymb}
\usepackage[utf8]{inputenc} 
\usepackage{colortbl}
\usepackage{pifont}
\usepackage{wrapfig}
\usepackage{array}

\definecolor{darkgreen}{rgb}{0,0.8,0} 

\usepackage[T1]{fontenc}

\usepackage[utf8]{inputenc}

\usepackage{microtype}

\usepackage{inconsolata}

\usepackage{graphicx}

\title{ActiView: Evaluating Active Perception Ability for Multimodal Large Language Models}

\author{Ziyue Wang\textsuperscript{*,1}, Chi Chen\textsuperscript{*,1},  Fuwen Luo\textsuperscript{*,1}, Yurui Dong\textsuperscript{3}, \\
{\bf Yuanchi Zhang\textsuperscript{1}, Yuzhuang Xu\textsuperscript{1}, Xiaolong Wang\textsuperscript{1}, Peng Li\textsuperscript{2}, Yang Liu\textsuperscript{1,2}} \\
  \textsuperscript{1}Dept. of Comp. Sci. \& Tech., Institute for AI, Tsinghua University, Beijing, China \\
  \textsuperscript{2}Institute for AI Industry Research (AIR), Tsinghua University, Beijing, China \\
  \textsuperscript{3}School of Management, Fudan University, Shanghai, China
  }

\DefineFNsymbols*{1}{*}
\setfnsymbol{1}

\begin{document}
\maketitle
\begin{abstract}
Active perception, a crucial human capability, involves setting a goal based on the current understanding of the environment and performing actions to achieve that goal. Despite significant efforts in evaluating Multimodal Large Language Models (MLLMs), active perception has been largely overlooked. To address this gap, we propose a novel benchmark named ActiView to evaluate active perception in MLLMs. 
We focus on a specialized form of Visual Question Answering (VQA) that eases and quantifies the evaluation yet challenging for existing MLLMs. Meanwhile, intermediate reasoning behaviors of models are also discussed. 
Given an image, we restrict the perceptual field of a model, requiring it to actively zoom or shift its perceptual field based on reasoning to answer the question successfully. We conduct extensive evaluation over 30 models, including proprietary and open-source models, and observe that restricted perceptual fields play a significant role in enabling active perception.
Results reveal a significant gap in the active perception capability of MLLMs, indicating that this area deserves more attention.
We hope that ActiView could help develop methods for MLLMs to understand multimodal inputs in more natural and holistic ways.\footnote{Codes and data will be available at \url{https://github.com/THUNLP-MT/ActiView}}
\end{abstract}

\section{Introduction}

The advent of Multimodal Large Language Models (MLLMs) has marked a significant milestone in the realm of artificial intelligence, demonstrating capabilities that are increasingly approaching human-like performance~\citep{OpenAI2023Gpt4v, liu2023visual, ye2024mplug}. This advancement, while promising, also presents new challenges and opportunities for evaluating these models. As a result, the landscape of MLLM evaluation is rapidly evolving, with numerous benchmarks being developed to either comprehensively evaluate models~\citep{fu2023mme, liu2023mmbench} or to analyze specific aspects of their capabilities~\citep{liu2023hallusionbench,lu2023mathvista, luo2024codis, xiao2024logicvista, li2024seed, nie2024mmrel, qian2024mia}.

\begin{figure}
    \centering
    \includegraphics[width=.48\textwidth]{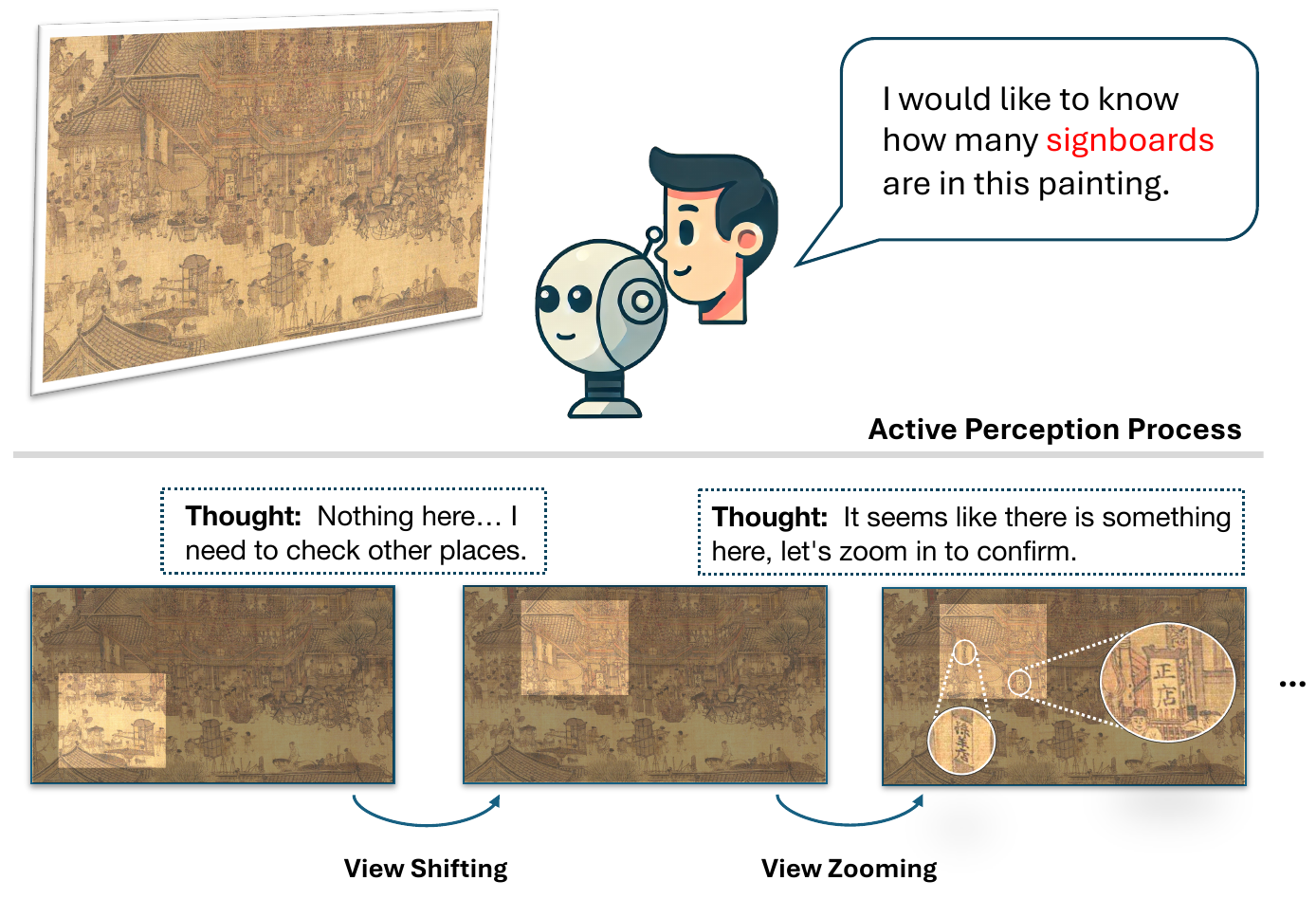}
    \vspace{-2em}
    \caption{Active perception allows humans or models to perform more complex tasks by actively seeking and processing relevant information. In this paper, we evaluate two key active perception abilities for MLLMs: 1) \textit{shifting}, as real-world scenarios often present limited views and require shifts to obtain new perspectives, and 2) \textit{zooming}, which helps enhance perception by zooming out for a broader view and zooming in for details.}\vspace{-1.6em}
    \label{fig:figure1}
\end{figure}

\begin{table*}[!t]
\centering\scriptsize
\begin{tabular}{@{\hspace{0.15cm}}c@{\hspace{0.25cm}}l@{\hspace{0.2cm}}|l@{\hspace{0.2cm}}|cc|cc|c}\toprule
& \multirow{2}{*}[-0.7ex]{Benchmarks} & \multirow{2}{*}[-0.7ex]{Evaluation Target} & \multicolumn{2}{c|}{Change of Per. Fields} & \multirow{2}{*}[-0.7ex]{\begin{tabular}[c]{@{}c@{}}Num.\\Img\end{tabular}} & \multirow{2}{*}[-0.7ex]{\begin{tabular}[c]{@{}c@{}}Evaluation\\Instances\end{tabular}} & \multirow{2}{*}[-0.7ex]{Annotator}  \\ \cmidrule(lr){4-5} 
 &  &  & Shifting & Zooming &  &  \\ \midrule
\multirow{6}{*}{\rotatebox{90}{\scriptsize{General}}} & MME~\citep{fu2023mme} & Visual comprehension  & \textcolor{red}{\ding{55}} & \textcolor{red}{\ding{55}}  & 1.1k & 1.3k & Manual \\
& MMBench~\citep{liu2023mmbench} & Visual comprehension   & \textcolor{red}{\ding{55}} & \textcolor{red}{\ding{55}}  & 1.8k & 1.8k & Manual + Auto  \\
& MM-Vet~\citep{yu2023mm} & Integrated capabilities  & \textcolor{red}{\ding{55}} & \textcolor{red}{\ding{55}}  & 200  & 218 & Manual* \\
& Seed-Bench~\citep{li2023seedbench} & Visual  comprehension  & \textcolor{red}{\ding{55}} & \textcolor{red}{\ding{55}} & 1.9k* & 24k & Auto   \\ 
& BLINK~\citep{fu2024blink} & Visual perception  & \textcolor{darkgreen}{\ding{51}} & \textcolor{red}{\ding{55}} & 7.3k & 3.8k & Manual \\
\midrule
\multirow{5}{*}{\rotatebox{90}{\scriptsize{Specialized}}} & ViP-Bench~\citep{biernacki2021vip} &  Understanding of visual prompt & \textcolor{red}{\ding{55}} & \textcolor{red}{\ding{55}} & 303 & 303 & Manual \\
& HallusionBench~\citep{liu2023hallusionbench} & Hallucination & \textcolor{red}{\ding{55}} & \textcolor{red}{\ding{55}} & 346 & 1.1k & Manual + Auto \\
& LogicVista~\citep{xiao2024logicvista} & Visual logical reasoning & \textcolor{red}{\ding{55}} & \textcolor{red}{\ding{55}} & 448 & 448 & Manual \\
& CNT~\citep{roberts2023charting} & Geographic and Geospatial & \textcolor{red}{\ding{55}} & \textcolor{darkgreen}{\ding{51}} & 345 & 345 & Manual\\
& V*~\citep{wu2023textit} & Fine-grained visual search & \textcolor{red}{\ding{55}} & \textcolor{darkgreen}{\ding{51}} & 191 & 191 & Manual \\ 
\midrule
& ActiView (Ours) & Active perception & \textcolor{darkgreen}{\ding{51}} & \textcolor{darkgreen}{\ding{51}} & 314 & 1,625 & Manual \\
\bottomrule
\end{tabular}
\vspace{-1.2em}
\caption{Comparison with other benchmarks for MLLMs. ``Per. Fields'': Perceptual Fields. 1.9k*: Videos. Manual*: A mixture of manual annotation and data from existing benchmarks.
Our benchmark focuses on evaluating active perception abilities via changes in visual perceptual fields, including shifting for compensating missing information, zooming for fine-grained details in the current fields, and a combination of both to mimic real-world scenarios.} 
\label{tab:compare}
\vspace{-2.2em}
\end{table*}

Despite the extensive efforts devoted to MLLM evaluation, \emph{active perception}~\citep{bajcsy1988active, bajcsy2018revisiting} remains underexplored. Active perception involves understanding the reasons for sensing, choosing what to perceive, and determining the methods, timing, and locations for achieving that perception~\citep{bajcsy2018revisiting}. This is important because in the real world, the desired information often does not appear directly in the center of one's field of vision. Instead, it requires individuals to move their field of view, locate details, and filter out distracting information. For example, in Figure~\ref{fig:figure1}, suppose we are looking for information in a giant painting. We need to first shift our view to locate the specific area and then possibly zoom in to gather detailed information.
Intuitively, active perception not only enables a person or model to accomplish more complex tasks, but also has the potential to serve as a good indicator of the level of intelligence of a model, making it a critical capability that warrants thorough evaluation.

\vspace{-2pt}
However, existing multimodal evaluation benchmarks are not well-suited to assess active perception capabilities. Table~\ref{tab:compare} summarizes several widely used or recently proposed multimodal evaluation benchmarks, most of them assess models in static perceptual field settings, where models process information presented directly to them without requiring active exploration or dynamic adjustments to their field of view. BLINK~\citep{fu2024blink}, V*~\citep{wu2023textit}, and CNT~\citep{roberts2023charting} are exceptions, as they utilize dynamic perceptual fields. However, they only consider either shifting or zooming of the field of view in specific scenarios, which are insufficient for measuring active perception capabilities. Therefore, there is a clear need for new evaluation frameworks that can adequately capture active perception abilities across diverse and dynamic environments. 

\vspace{-2pt}
To fill this gap, we introduce a novel benchmark specifically designed to evaluate \textbf{Acti}ve perception through \textbf{View} changes~(\textbf{ActiView}). Given the difficulty of comprehensively evaluating such capabilities across all possible scenarios, ActiView focuses on a series of tasks that are feasible to evaluate, yet still present significant challenges to current models. We manually curate a diverse set of instances, each including question-answer pairs and reasoning clues, and follows the Visual Question Answering (VQA)~\citep{antol2015vqa} format but exhibits additional features: 1) Each question requires an understanding of multiple detailed visual clues in the image to answer accurately. 2) View constraints are imposed, allowing models to perceive only a partial field of view of the full image at a time. This setup explicitly requires models to perform view shifting and zooming to gather necessary information and eliminate potential distractions, simulating the active perception process in real life. 3) In addition to answering visual questions, intermediate reasoning behaviors, such as view selection, also contribute to the evaluation. 

Results from over 30 models reveal that these models generally lag behind in active perception. For instance, the strong proprietary model, GPT-4o, only achieved an average score of 66.40\% with our designed evaluation pipelines for fundamental abilities, which is notably lower then the human score of 84.67\%. Regarding another pipeline that allows models to flexibly integrate these fundamental abilities, GPT-4o achieves 69.54\%, implying that combining fundamental active perception abilities can contribute to improvements. Moreover, the average performance gap between proprietary models and open-source models in active perception is considerably smaller within our designed pipelines than those observed in tasks from previous research. Recent small open-source models, in particular, exhibit approaching GPT-4o results. Experimental results suggest that models tend to perform better when given a complete image but struggle to develop a holistic understanding when presented with even all the separate and constrained perceptual fields. These findings highlight the need for further research in active perception and the value of our benchmark for advancing this field.

\begin{figure*}[t]
    \centering
    \includegraphics[width=.99\textwidth]{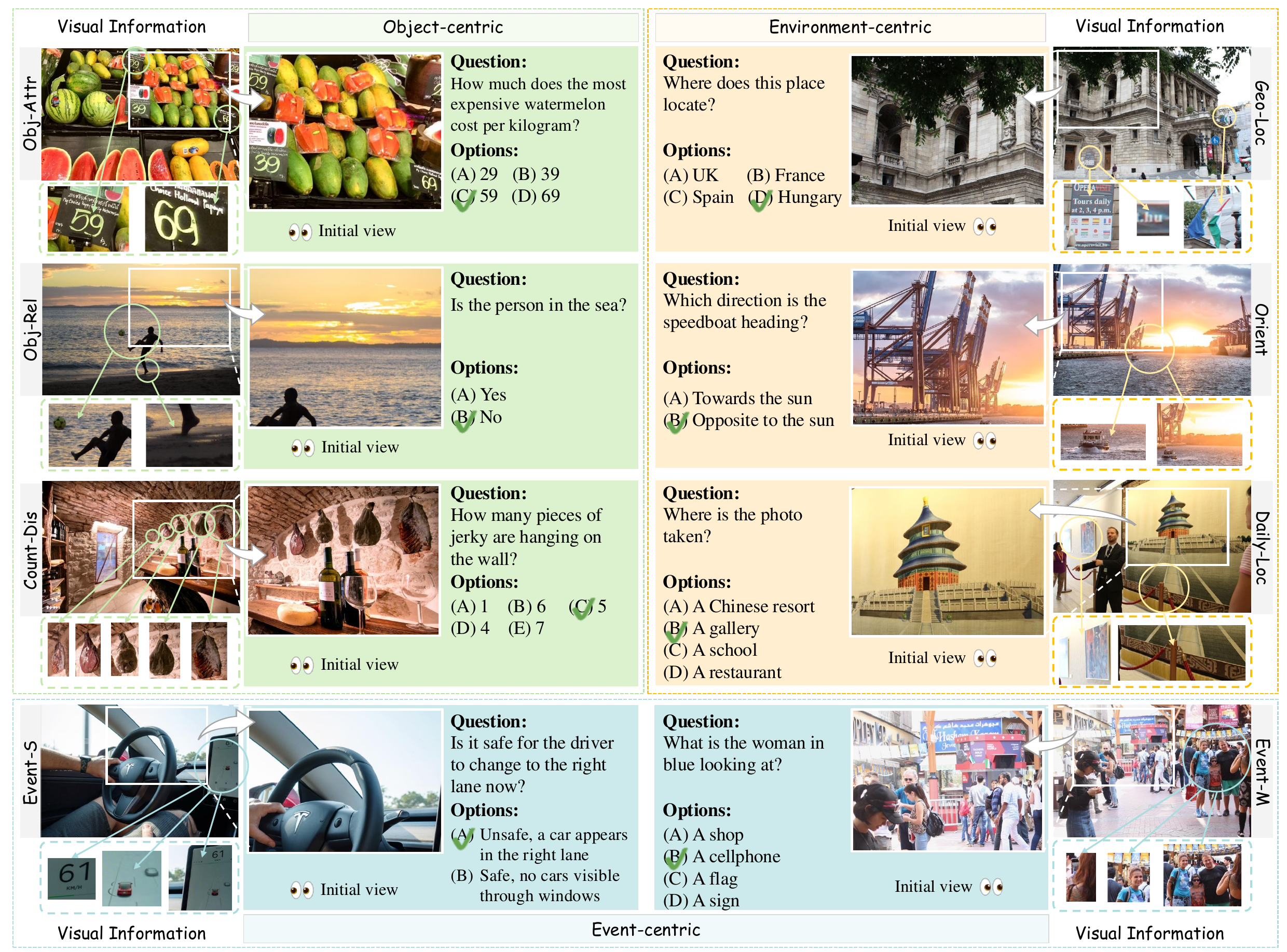}
    \vspace{-0.7em}
    \caption{Examples of ActiView, exhibiting the following features: i) requiring focusing on multiple fine-grained regions; ii) requiring distinguishing distracting information from the entire image; iii) requiring moving of perceptual fields to obtain sufficient visual information to answer questions. During evaluation, models will be given an initial view cropped from the original image as shown above. Visual Information: human-annotated visual clues.\label{fig:example}}
    \vspace{-1.4em}
\end{figure*}

\section{Related Works}

\subsection{MLLM Benchmarks}
Extensive efforts have been devoted to developing MLLM evaluation benchmarks (Table~\ref{tab:compare}). covering a wide range of capabilities, including visual comprehension ~\citep{fu2023mme,liu2023mmbench, fu2024video}, visual perception~\citep{fu2024blink}, hallucination~\citep{liu2023hallusionbench}, and mathematical and logical reasoning~\citep{lu2023mathvista, xiao2024logicvista}. However, most of them rely on a static view of the input image, which is not suited for assessing active perception. While BLINK~\citep{fu2024blink} involves view shifting, and both V*~\citep{wu2023textit} and CNT~\citep{roberts2023charting} require view zooming, active perception is not a prerequisite for solving their evaluation questions, making them insufficient for comprehensive active perception evaluation. In contrast, our benchmark considers both view shifting and zooming, with questions specifically designed to necessitate active perception for answering, which makes it a more robust framework towards active perception evaluation. 

\vspace{-0.5em}
\subsection{Active Perception in MLLMs}
Although MLLMs have attracted extensive interest, less effort has been dedicated to improving the active perception capability of MLLMs.
One line of research focuses on improving the ability of processing high-resolution images by using higher-resolution ViTs~\citep{ye2024mplug}, slicing high-resolution images and then concatenate them~\citep{liu2024llavanext}, or directly using LLMs to process raw patches of any resolution~\citep{li2023otterhd}. The other line emphasizes visual search for fine-grained details. SEAL~\citep{wu2023textit} fine-tunes a framework of two MLLMs to follow the visual search mechanism for precise visual grounding, and V-IRL~\citep{yang2024virl} proposes an active detection strategy to improve the comprehension of real-world geospatial information.
Despite these efforts, our evaluation results reveal that existing MLLMs still generally lack active perception capabilities. Our benchmark will shed light on evaluating and enhancing active perception in MLLMs.

\section{ActiView}
\vspace{-0.2em}
Our benchmark examines active perception abilities through different perceptual fields, where \textbf{Acti}vely zooming and shifting of \textbf{View}s (\textbf{ActiView}) are required. We summarize \textit{zooming} and \textit{shifting} as core components of active perception, as depicted in Figure~\ref{fig:figure1}, which allow us to evaluate active perception abilities of models both separately and integratedly. ActiView imitates the behavior of active perception by providing models with a constraint initial view, either a cropped field of the original image or a full image at reduced resolution. As shown in Figure~\ref{fig:example}, models should search for missing critical information through view zooming and shifting, while eliminating distractions caused by redundant content within the view. 

\vspace{-0.5em}
\subsection{Benchmark Overview}\label{sec:design}
When perceiving an image, humans instinctively focus on three principle aspects: the depicted environment, the primary objects, and the events in which these objects are involved. Similarly, we categorize questions in our benchmark into three main types, as shown in Figure~\ref{fig:example}, which are further divided into eight sub-classes according to the type of visual information and features required to answer the questions. Due to page limitations, detailed descriptions and typical examples for each sub-class are provided in Appendix~\ref{app:detail_cate}. Below is a concise overview of categories in ActiView:
\vspace{-6pt}
\begin{itemize}[left=0.3cm, itemsep=0pt, parsep=0pt]
    \item \textbf{Environment-centric (Type I)} involves three sub-classes. \textit{Geo-Localization} (\textit{Geo-Loc}) focuses on geographical features unique to specific countries or cities, requiring models to identify geographical locations implied in the image. \textit{Orientation} (\textit{Orient}) challenges models to reason from information of natural orientation. \textit{Daily-location} (\textit{Daily-Loc}) distincts from Geo-Loc by centering on everyday locations that could appear in most cities, without being tied to a particular country or city.
    \item \textbf{Object-centric (Type II)} tasks go beyond simple grounding that directly ask for the attributes or relations of objects, by challenging models with distracting information. Three sub-classes are \textit{Object-attribute} (\textit{Obj-Attr}), that focuses on identifying objects attributes out of distractions that potentially mislead the model; \textit{Object-relation} (\textit{Obj-Rel}), that examines spatial relationships among objects while the questions do not explicitly ask for them; and \textit{Counting} (\textit{Count-Dis}), that involves counting objects while handling similar but distracting elements that can lead to incorrect answers.
    \item \textbf{Event-centric (Type III)} focuses on the interactions between humans and objects, such as actions and activities. This category is divided according to the number of objects involved in the target event, including \textit{Event-single} (\textit{Event-S}) which focuses on events involving a single item or person; and \textit{Event-multi} (\textit{Event-M}) that happens among multiple items or people. 
\end{itemize}
\vspace{-6pt}

\begin{figure}
    \centering
    \includegraphics[width=.95\columnwidth]{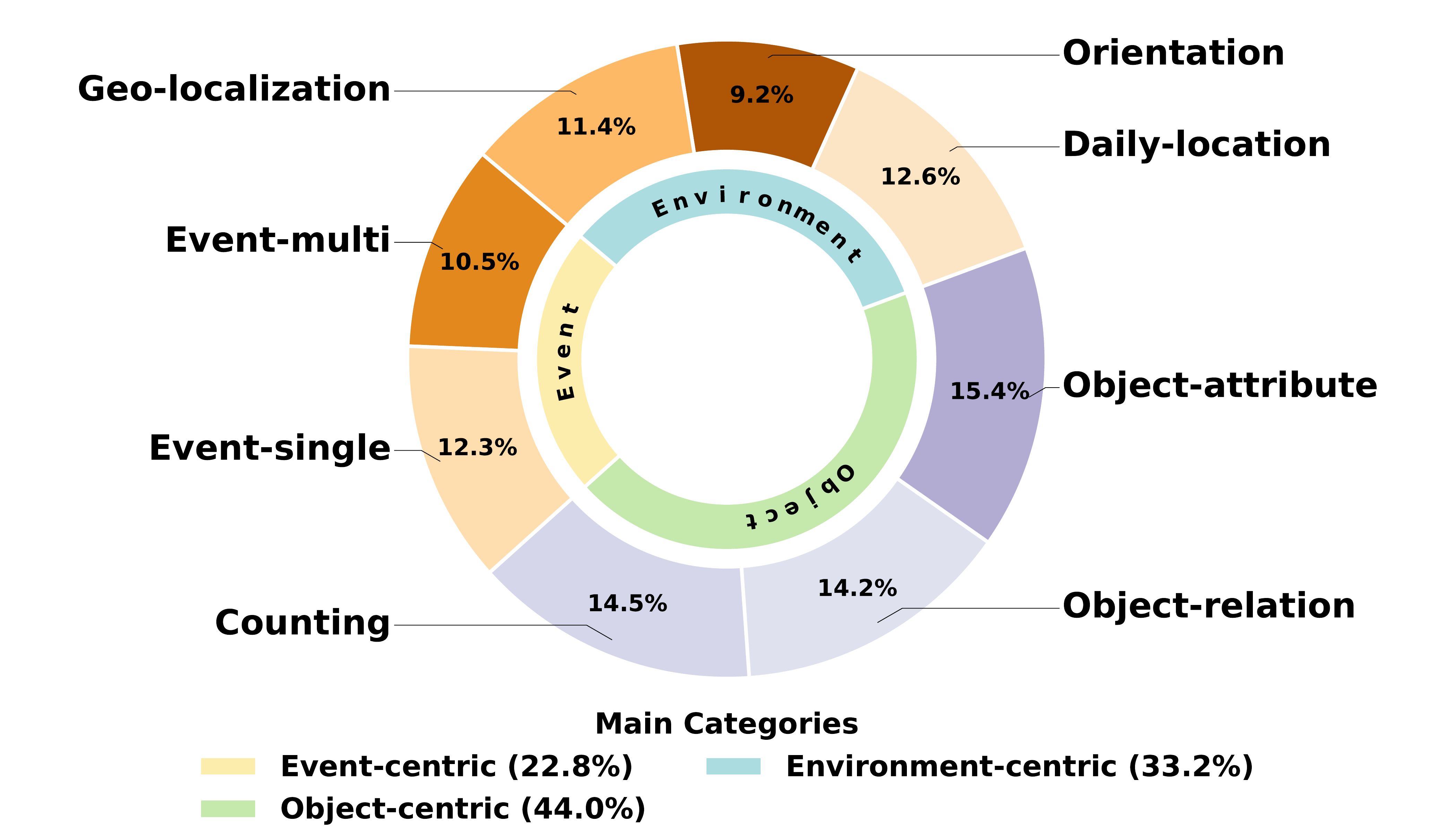}
    \vspace{-0.8em}
    \caption{The statistical distribution of our benchmark.\label{fig:dist}}
    \vspace{-1.5em}
\end{figure}

\begin{figure*}[t]
    \centering
    \includegraphics[width=.97\textwidth]{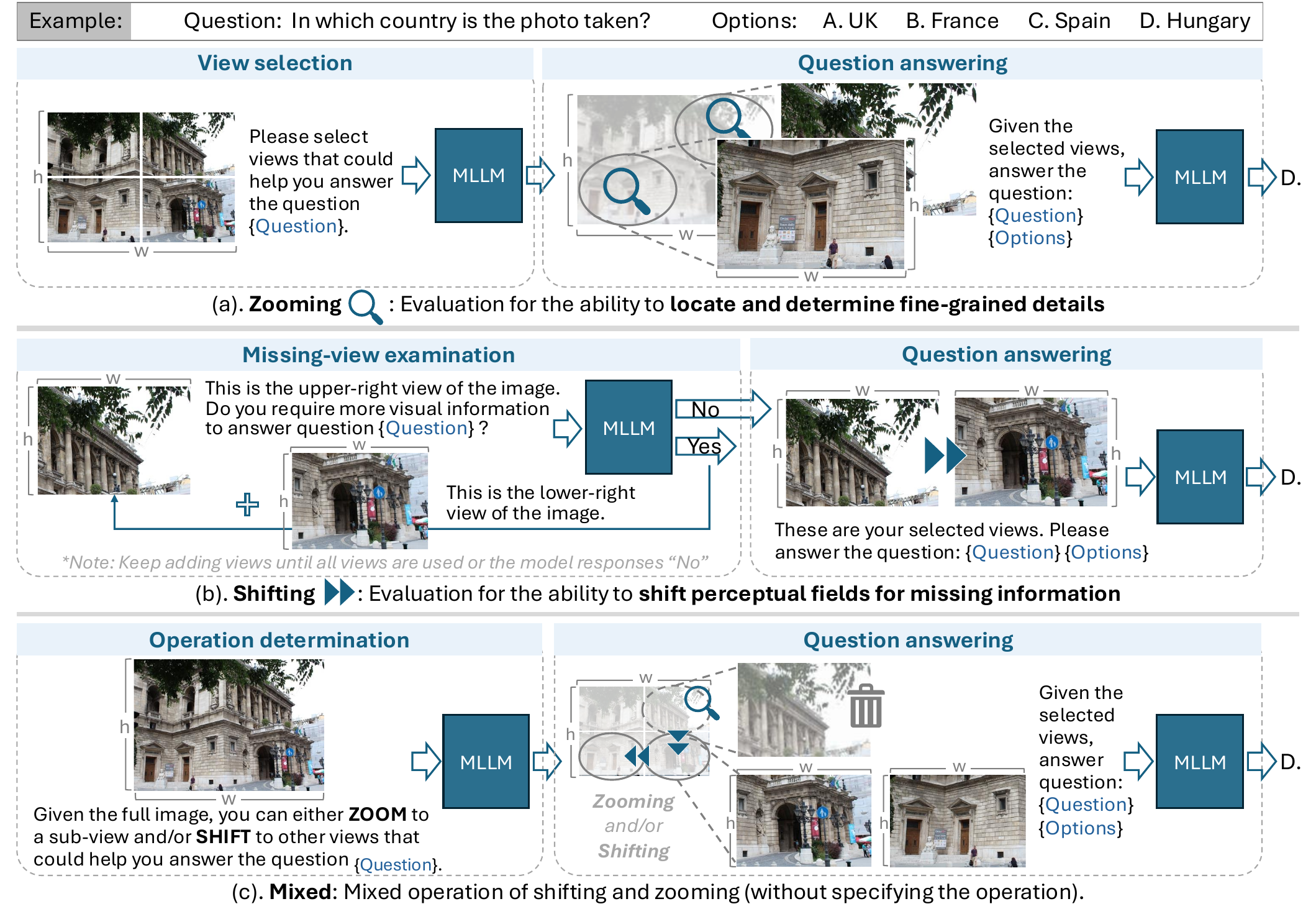}
    \vspace{-0.5em}
    \caption{Evaluation pipelines as described in \S\ref{sec:pipeline}. (a) \textbf{Zooming} requires models to select multiple regions to zoom in. It tests one of the fundamental active perception abilities. (b) \textbf{Shifting} challenges models to ask for more necessary information. It tests the other fundamental active perception abilities. (c) \textbf{Mixed} simulates human behavior when shifting perceptual fields for missing information. It is more flexible and applicable in real life compare to the previous two fundamental abilities. Note that while we provide an example in the figure where model delete a zoomed sub-view, the deletion behavior is NOT required. It is to address the compound features of the mixed pipeline (c) compare to the other fundamental pipelines (a) and (b).}\label{fig:pipeline}
    \vspace{-1em}
\end{figure*}

\vspace{-0.5em}
\subsection{Data Curation and Statistics} \label{sec:collect}
\vspace{-0.3em}
Our dataset is manually curated, including image collection, question and option annotation, and visual clue identification. To assess active perception abilities, which require zooming for fine-grained details and shifting for capturing missing information, we select images featuring multiple fine-grained objects and complex scenes or events. Detailed descriptions of the collected images and annotation guidelines can be found in Appendix~\ref{app:data}. Annotators are also required to identify visual clues to support their answers, as shown in the ``Visual Information'' columns in Figure~\ref{fig:example}. 
To prevent models from selecting answers based solely on the provided options, we adopt a set of more flexible annotation rules than the typical two- or four-option format. Option count ranges from two to seven, many of which are derived directly from the images. For instance, in the Geo-Loc task shown in Figure~\ref{fig:example}, the options all correspond to visual clues in the image, such as flags representing the UK, France, Spain, and Hungary. Furthermore, for options comprised by numbers, they are arranged in random order to avoid biased predictions.

The experiments of automatic data generation are discussed in Appendix~\ref{app:auto_gen}. In summary, we found that powerful models, such as GPT-4V and GPT-4o, fail to satisfy our annotation guidelines. They struggle with hallucination when processing multiple images, and fall short on distinguish between visual facts in the image and external world knowledge not present in the image.

\vspace{-6pt}
\paragraph{Statistics.} We collected 314 images and annotated 325 questions, with distribution of categories shown in Figure~\ref{fig:dist}. To further enrich the diversity, each question corresponds to 5 different evaluation instances, assessing active perception across varying components and difficulties. In total, 1,625 evaluation instances are curated. On average, there are 3.24 options and 2.64 sub-views containing visual clues per question, highlighting that a single view is often insufficient for answering accurately, and that the ability to comprehend multiple images jointly is crucial for our benchmark. 

\vspace{-0.2em}
\section{Evaluation}\label{sec:eval}
\vspace{-0.3em}
For thorough investigation, we design three evaluation pipelines for different operations of perceptual fields as illustrated in Figure~\ref{fig:pipeline}, including two individual pipelines for core components, and a mixed pipeline incorporating both. We set up five different initial views for each question-image pairs, where a full image of limited resolution is used for zooming and mixed pipelines, and four constrained views are applied for the shifting pipeline. These correspond to the 1,625 evaluation instances mentioned in previous section. Due to page limitation, details of pipelines can be found in Appendix~\ref{sec:pipeline}, and evaluated models are discussed in Appendix~\ref{app:model}.

\begin{table*}[t]
\centering
\resizebox{\textwidth}{!}{
\begin{tabular}{cr|ccc|ccccccccc|ccccc}
\toprule
 & \multirow{2}{*}[-1.8ex]{Metrics} & \multicolumn{3}{@{\hspace{0.05cm}}c@{\hspace{0.01cm}}|@{\hspace{0.05cm}}}{Proprietary Models}  & \multicolumn{9}{@{\hspace{0.05cm}}c@{\hspace{0.01cm}}|@{\hspace{0.05cm}}}{Multi-image Open-source Models} & \multicolumn{5}{@{\hspace{0.05cm}}c@{\hspace{0cm}}}{Single-image Open-source Models} \\ \cmidrule(lr){3-5} \cmidrule(lr){6-14} \cmidrule(lr){15-19} \vspace{-2pt}
&  & {\begin{tabular}[c]{@{}c@{}}Gemini\\1.5-pro\end{tabular}} & GPT-4o & {\begin{tabular}[c]{@{}c@{}}Claude3.5\\Sonnet\end{tabular}} & {\begin{tabular}[c]{@{}c@{}}Qwen2.5\\VL-7B\end{tabular}} & {\begin{tabular}[c]{@{}c@{}}DeepSeek\\-VL2\end{tabular}} & {\begin{tabular}[c]{@{}c@{}}Idefics\\-3-8B\end{tabular}} & {\begin{tabular}[c]{@{}c@{}}MiniCPM\\-V 2.6\end{tabular}} & {\begin{tabular}[c]{@{}c@{}}mPLUG-\\Owl3-7B\end{tabular}} & {\begin{tabular}[c]{@{}c@{}}LLaVA\\-OV\end{tabular}} & {\begin{tabular}[c]{@{}c@{}}Intern\\VL2-8B\end{tabular}} & {\begin{tabular}[c]{@{}c@{}}Mantis\\-8B\end{tabular}} & {\begin{tabular}[c]{@{}c@{}}Phi-3.5\\-vision\end{tabular}} & {\begin{tabular}[c]{@{}c@{}}GLM-\\4V-9B\end{tabular}} & {\begin{tabular}[c]{@{}c@{}}InternVL\\13B\end{tabular}} &  {\begin{tabular}[c]{@{}c@{}}LLaVA\\-1.6 7B\end{tabular}} & {\begin{tabular}[c]{@{}c@{}}MGM\\-7B-HD\end{tabular}} & SEAL \\
\midrule
\multirow{4}{*}{\rotatebox{90}{\scriptsize{Zooming}}} & $P_{\text{select}}$ & 79.00 & 79.60 & \textbf{81.49} & \colorbox{gray!10}{73.72} & 71.64 & 71.31	& 69.13	& 68.25 & 69.08 & 71.03 & 64.88 & 69.60 & \colorbox{gray!10}{74.93} & 70.48 & 65.46 & 66.24 & 65.92 \\
& $R_{\text{select}}$ & 62.63 & \colorbox{gray!10}{69.03} & 67.64 & \textbf{70.55} & 55.30 & 41.09 & 57.03 & \colorbox{gray!10}{68.57} & 46.47 & 41.09 & 22.80 & 28.52 & 30.62 & 61.70 & 68.57 & 30.15 & 68.22\\
& $F_1$ & 69.87 & \textbf{73.94} & 73.92 & \colorbox{gray!10}{72.10} & 62.42 & 52.14 & 62.50 & 68.41 & 47.91 & 52.06 & 33.74 & 40.46 & 43.47 & 65.80 & 66.98 & 41.44 & \colorbox{gray!10}{67.05} \\
& ACC$_{\text{QA}}$ & \textbf{72.31} & 68.62 & 71.69 & \colorbox{gray!10}{68.92} & 65.85 & 58.15 & 61.85 & 60.92 & 65.23 & 56.00 & 60.62 & 56.62 & 56.92 & 62.77 & \colorbox{gray!10}{68.92} & 34.77 & 54.77 \\
\midrule
\multirow{4}{*}{\rotatebox{90}{\scriptsize{Shifting}}} & ACC$_{\text{Shift-R}}$ & \colorbox{gray!10}{67.08} & \colorbox{gray!10}{67.08} & 65.23 & \textbf{68.62} & 65.54 & 61.85 & 54.77 & 51.69 & 53.54 & 54.77 & 52.92 & 50.46 & \colorbox{gray!10}{56.92} & 53.85 & 51.69 & 48.62 & 42.77 \\
& ACC$_{\text{Shift-E}}$ & \textbf{67.38} & 66.77 & 66.15 & \colorbox{gray!10}{67.08} & 62.77 & 59.83 & 61.23 & 56.31 & 57.23 & 59.70 & 55.38 & 54.15 & \colorbox{gray!10}{60.62} & 52.92 & 52.31 & 48.00 & 42.77 \\
& ACC$_{\text{Shift-M}}$ & \colorbox{gray!10}{65.54} & 65.23 & 60.31 & \colorbox{gray!10}{67.38} & 64.31 & 59.69 & 58.15 & 55.69 & 52.31 & 53.23 & 52.92 & 50.15 & \colorbox{gray!10}{56.00} & 52.92 & 49.32 & 47.69 & 40.02 \\
& ACC$_{\text{Shift-H}}$ & \colorbox{gray!10}{67.69} & 64.31 & 61.85 & \textbf{68.00} & 64.62 & 60.31 & 55.69 & 53.54 & 48.62 & 52.00 & 52.31 & 45.54 & \colorbox{gray!10}{52.92} & 51.08 & 48.00 & 50.15 & 40.62 \\
\midrule
\multicolumn{2}{@{\hspace{0cm}}c@{\hspace{0.05cm}}|@{\hspace{0.05cm}}}{Average ACC} & \textbf{68.00} & 66.40 & 65.05 & \textbf{68.00} & 65.11 & 59.88 & 58.34 & 55.63 & 55.39 & 55.14 & 54.83 & 51.38 &  \colorbox{gray!10}{56.68} & 54.71 & 54.03 & 45.84 & 44.07 \\
\midrule
\multicolumn{2}{@{\hspace{0cm}}c@{\hspace{0.05cm}}|@{\hspace{0.05cm}}}{\begin{tabular}[c]{@{}c@{}}ACC w/\\Human Clues\end{tabular}} & 72.00 & \colorbox{gray!10}{73.54} & 72.31 & 70.77 & 68.31 & 60.92 & 62.77 & 60.62 & 64.92 & \colorbox{gray!10}{73.23} & 59.38 & 58.15 & \textbf{74.46} & 68.00 & 67.69 & 68.31 & 56.92 \\
\bottomrule
\end{tabular}
}
\vspace{-0.5em}
\caption{Results of evaluation of individual components, following shifting and zooming pipelines. We list results of some widely-discussed models here, and refer readers to Table~\ref{tab:full_rst} for more details. The human performance is \textbf{84.67\%} referring to Table~\ref{tab:human_detail} in Appendix~\ref{app:human}. ``Average AVG'': average scores of question answering accuracy of all settings. The best scores of each row are \textbf{bolded} and the best scores in the other model types are \colorbox{gray!10}{highlighted}.} \label{tab:rst_}
\vspace{-1em}
\end{table*}

\begin{table*}[t]
        \centering\scriptsize
        \begin{tabular}{@{\hspace{0.1cm}}l|@{\hspace{0.1cm}}c@{\hspace{0.1cm}}c@{\hspace{0.1cm}}c@{\hspace{0.1cm}}|@{\hspace{0.1cm}}c@{\hspace{0.1cm}}c@{\hspace{0.1cm}}c@{\hspace{0.1cm}}c@{\hspace{0.1cm}}c@{\hspace{0.1cm}}c@{\hspace{0.1cm}}c}
        \toprule
         \multirow{2}{*}[-1.8ex]{Metrics} & \multicolumn{3}{c|@{\hspace{0.1cm}}}{Proprietary Models}  & \multicolumn{7}{c}{Multi-image Open-source Models}\\ \cmidrule(lr){2-4} \cmidrule(lr){5-11}\vspace{-3pt}
         & {\begin{tabular}[c]{@{}c@{}}Claude 3.5\\Sonnet\end{tabular}} & GPT-4o & Gemini-1.5-pro & {\begin{tabular}[c]{@{}c@{}}Qwen2.5-VL\\-7B\end{tabular}} & {\begin{tabular}[c]{@{}c@{}}Qwen2-VL\\-7B\end{tabular}} & DeepSeek-VL2 & {\begin{tabular}[c]{@{}c@{}}Qwen2.5-VL\\-3B\end{tabular}} & {\begin{tabular}[c]{@{}c@{}}MiniCPM\\-V 2.6\end{tabular}} & {\begin{tabular}[c]{@{}c@{}}Idefics3\\-8B\end{tabular}} & {\begin{tabular}[c]{@{}c@{}}mPLUG-Owl3\\-7B\end{tabular}}\\ 
        \midrule
        \#zoom & 2.30 & 1.61 & 1.82 & 1.88 & 2.51 & \colorbox{blue!12}{2.65} & 1.21 & 1.31 & \colorbox{yellow!12}{1.16} & 2.59 \\
        \#shift & \colorbox{blue!12}{3.15} & 1.23 & 1.65 & 1.59 & 2.17 & 1.74 & 1.73 & \colorbox{yellow!12}{0.39} & 0.59 & 1.49 \\
        \#view (diff) & 1.89 (-0.75) & 1.35 (-1.29)  & 1.14 (-1.50) & 1.26 (-1.38) & \underline{2.12 (-0.52)} & \textbf{2.43 (-0.21)} & 0.94 (-1.70) & 0.94 (-1.70) & 0.58 (-2.06)  & 1.43 (-1.21)\\
        \midrule
        ACC & \textbf{72.00} & 69.54 & 68.92 & \underline{70.77} & 65.54 & 65.23 & 64.62 & 64.00 & 62.15 & 59.69 \\
        \bottomrule
        \end{tabular}
        \vspace{-1em}
        \caption{Experimental results of mixed pipeline for integrated components. ``\#zoom'': average zooming operations; ``\#shift'': average shifting operations. The \colorbox{blue!10}{most count} is marked in blue and the \colorbox{yellow!12}{least count} in yellow. ``\#view'': average used views; ``diff'': \#view-\#view\_annotated, indicating the difference between actual selected views and the average sub-views containing visual clues (2.64).  ``ACC'': question answering accuracy. The best results for \#view (diff) and ACC are \textbf{bolded} and the second best are \underline{underlined}.} \label{tab:mix_rst}
        \vspace{-1.5em}
\end{table*} 

\vspace{-0.2em}
\subsection{Pipelines for Individual Component} 
\vspace{-0.2em}
We separately investigate two core components, \textit{zooming} and \textit{shifting}.
The \textbf{zooming pipeline} evaluates the ability to locate and determine necessary fine-grained information. As shown in Figure~\ref{fig:pipeline} (a), this pipeline contains two stages, view selection and question answering. In this pipeline, models first select sub-views to zoom in given the initial view, the full image of size $w \times h$, then answer the question based on these zoomed views. The selected sub-views are resized to $w \times h$, the same as the initial view, to enable a zooming operation. Note that a ``None'' selection is permitted. 

The \textbf{shifting pipeline} addresses the ability to navigate perceptual fields incrementally, mimicking real-world scenarios where complete context is unavailable. It measures the ability to shift perceptual fields for missing information and to infer the answer jointly based on constrained perceptual fields. This is also a two-stage pipeline as in Figure~\ref{fig:pipeline} (b). To simulate the movement of human eyes, the model begins with an initial view of size $w \times h$, and determines if the current views are sufficient for answering the question. If more views are needed, adjacent views will present until the answer can be inferred.
Furthermore, we assign four difficulty levels according to human-annotated visual clues contained in the initial views, namely ``Shift-R'', ``Shift-E'', ``Shift-M'', and ``Shift-H'', with corresponding settings are detailed in Appendix~\ref{sec:pipeline}.

\subsection{Pipeline for Integrated Components} 
In addition to the individual pipelines, we also implement an automated mixed setting, \textbf{mixed pipeline}, that does not specify the type of active perception ability required. As illustrated in Figure~\ref{fig:pipeline} (c), models must independently decide whether to zoom, shift, or use both to address different perceptual fields. In contrast to the zooming pipeline, where models answer questions based on all selected views, the mixed pipeline allows models to discard irrelevant views after selection. Unlike the shifting pipeline, the mixed pipeline also provides access to the full image view in addition to cropped sub-views. 
The mixed pipeline emphasizes model autonomy and requires models to account for all views, including the full one, to ensure unbiased operation decision and view selection. 
Otherwise, it is at risk of reverting to either zooming or shifting tasks, limiting its evaluation. The need for autonomy, strict adherence to instructions, and comprehensive understanding of all views makes this pipeline suitable only for the most advanced and recent multi-image models.

Templates for the above pipelines are listed in Appendix~\ref{app:template_zoom}, ~\ref{app:template_shift} and ~\ref{app:template_mix}, respectively. In addition, we enable a general VQA evaluation, where models answer visual questions given full images without zooming or shifting, to serve as the reference for assessing the difficulty of our created benchmark. Detailed prompts are provided in Appendix~\ref{app:template_general}.

\vspace{-0.5em}
\subsection{Processing of Views} \label{sec:process}
In this paper, we focus on the interleaved multi-image setting, which is more practical and natural than the single-image setting. Multi-image models can naturally comprehend several views at one time during evaluating, allowing us to directly format the images and text in interleaved form. For fairness, we also design evaluation methods for single-image models.
To enable zooming and shifting operations, images are split into four sub-views, ensuring unbiased evaluation across all models, regardless of their training data or image processing strategies. We also discuss different splitting methods (Appendix~\ref{app:split}), and input view processing techniques, including image processing (Appendix~\ref{app:process}) and textual form conversion (Appendix~\ref{app:caption}).

\section{Results and Analysis} 

\begin{table}[t]
\centering\small
\begin{tabular}{@{\hspace{0.1cm}}c|c@{\hspace{0.2cm}}c@{\hspace{0.1cm}}|c@{\hspace{0.1cm}}}
\toprule
- & Human* & Random & Text-only (GPT-4o)\\
\midrule
ACC & 84.67 & 33.95 & 2.45 \\
\bottomrule
\end{tabular}
\vspace{-1em}
\caption{Human level performances, random choice result (averaged over 10k runs), and text-only evaluation. Detailed discussion of human evaluation is provided in  Appendix~\ref{app:human}, and text-only evaluation across different models is provided in Appendix~\ref{app:textonly}.} 
\label{tab:human_rst}
\vspace{-1.5em}
\end{table}

Experimental results of individual components, zooming and shifting, are in listed Table~\ref{tab:rst_}. We adopt accuracy as the evaluation metric for question answering, together with measurements of view selection (detailed in Appendix~\ref{app:view_measure}). Due to space constraints, we only list selected high-performance models in these two tables, and provide elaborated results of all evaluated models in Appendix~\ref{app:more_rst}, including scores for each categories, sub-classes and difficulties. Experimental results of integrated components from the mixed pipeline are reported in Table~\ref{tab:mix_rst}. 
The number of candidate options ranges from two to seven, with a random choice baseline of 33.95\% on our benchmark. We also conduct human and text-only evaluation to assess the difficulty and robustness of our benchmark. The average performance among six testers is 84.67\%, indicating that while our benchmark is feasible for humans, it can still be challenging. The text-only evaluation implies that questions in ActiView cannot be solved solely by commonsense knowledge within models, and that visual information is crucial for completing the tasks.

\vspace{-0.5em}
\subsection{Main Results} 
\paragraph{Results of evaluation for individual components.} We draw four key findings from the pipelines for individual components of active perception. 

First, as shown in Table~\ref{tab:rst_}, all evaluated models outperform random guessing, indicating their potential to maintain active perception abilities of zooming and shifting. However, even the best proprietary models fall significantly behind human. Second, although proprietary models achieve better overall performances compared to open-source models, the performance gap between these two categories are considerably smaller compared to gaps observed in other tasks from previous research. Moreover, the gap is becoming smaller for some recent released open-source models such as Qwen2.5-VL, which achieves the same highest average score of 68.00\% as Gemini-1.5-pro. Third, among open-source models, multi-image models largely outperform single-image models, particularly in shifting evaluations with constrained views. Finally, we observe that the view selection scores, F$_1$ in particular, are highly related to the final performance. Lower scores of precision, recall and F$_1$ stand for more unnecessary information given, which also correlates with lower QA performances.

For results of mixed evaluation in Table~\ref{tab:mix_rst}, we observe that the evaluated models benefit from enabling complex active perception and often outperform individual zooming or shifting on average. Notably, MiniCPM-V 2.6 (64.00\%) and Idefics3-8B-Llama3 (62.15\%) surpass the accuracy of given human-annotated views (62.77\% and 60.92\%, respectively, from Table~\ref{tab:gt_rst}); and Claude 3.5 Sonnet and Qwen2.5-VL-7B achieve equivalent performance of given human-annotated visual clues. 

\paragraph{Results of evaluation for integrated components.} The mixed pipeline encourages models to zoom and/or shift perceptual fields autonomously, mimicking human behaviors and highlighting the effectiveness of active perception. However, during experiments, we noticed that some large multi-image models, typically released months or a year ago, failed to follow instructions in the mixed evaluation, generating irrelevant responses or selecting invalid views, disrupting the mixed process. In contrast, recent models, regardless of size, successfully handled the this evaluation. With the reported counts of operations, we can conclude that models that actively zoom and shift views are likely to present higher question answering scores, thereby exhibiting better active perception ability. 

\begin{table}[t]
\centering\scriptsize
\begin{tabular}{l|c|c|c}
\toprule
\textbf{Models} & \textbf{W/ Clues} & \textbf{Full} & \textbf{Zooming} \\
\midrule
GPT-4o & \textbf{73.54} & 67.38 & 68.62 \\
Qwen2-VL-7B & \textbf{65.85} & 63.08 & 64.62 \\
Qwen2.5-VL-3B & \textbf{66.15} & 65.85 & \textbf{66.15}  \\
Qwen2.5-VL-7B & \textbf{70.77} & 67.08 & 68.92  \\
InterVL2-8B & \textbf{73.23} & 58.15 & 56.00 \\
Qwen2-VL-7B & \textbf{65.85} & 63.08 & 64.62 \\
Idefics3-8B & \textbf{60.92} & 59.08 & 58.15 \\
SEAL & \textbf{56.92} & 48.31 & 54.77 \\
\bottomrule
\end{tabular}
\vspace{-1em}
\caption{Performance comparison among providing annotation clues (W/ Clues), full images without applying perceptual constraints (Full), and our designed zooming pipeline (Zooming).} \label{tab:full_gt}
\vspace{-1em}
\end{table}

\paragraph{Discussion of the necessity of active perception.} 
To address the effectiveness and significance of active perception, we conduct comparison among three settings: i) providing annotation clues, ii) full images without applying perceptual constraints, and iii) our designed zooming pipeline, to demonstrate the usefulness of active perception, and also highlight the current limitations of models in active perception. We select several high-performance models from our experiments, and report corresponding results in Table~\ref{tab:full_gt}, where active perception improves over Full for both manually guided scenarios (W/ Clues) and model automated scenarios (Zooming). Results suggest that if models could accurately identify the necessary views (such as obtaining the human-annotated clues), their performance could be further improved via active perception.
More case studies are further discussed in Appendix~\ref{app:case_study} to support the necessity of active perception.

\subsection{Analysis of Different Pipelines}
\paragraph{Impact of selected views}
Our pipelines involve selecting useful view in their first stages. The reliability of selected views plays a crucial role in the following question answering stage. We refer readers to Appendix~\ref{app:view_select} for elaborate discussion on corresponding results (in Table~\ref{tab:gt_rst}).

Overall, lower selection recall tends to correlate with lower VQA accuracy. For example, Idefics3-8B-Llama3 and InternVL2-8B present the lowest recalls (41.09\%) among multi-image models, leading to lower accuracies for zooming evaluation, 56.00\% and 58.15\%, respectively. We also investigate the performance when given groundtruth views that contain human-annotated clues. Generally, models are prompted to generate more accurate answers compared to the pure zooming setting. However, mPLUG-Owl3, Gemini-1.5-pro, and LLaVA-OneVision are only exceptional, whose performance slightly decrease when given visual clues. We argue that they are better at the question answering task rather than exhibiting active perception ability. Additionally, we observe that shifting evaluations tend to require more views for answering questions compared to zooming evaluation, yet often results in inferior overall performance, indicating that models lack the ability to actively shifting perceptual fields under constraints. Thus, we believe that more attention should be paid to evaluating and enhancing active perception abilities of MLLMs given limited perceptual fields. 

\paragraph{Performance for different difficulty levels}
Generally, the accuracy of question answering and the recall of view selection decrease as the difficulties of the initial views increases. As shown by typical results of LLaVA-OneVision and GLM-4V-9B in Table~\ref{tab:rst_}, the gaps between easy and hard settings are as large as 8.61\% and 7.70\%, respectively. 
However, exceptions exist for Gemini-1.5-pro and Idefics3, demonstrating different reasons, where one is caused by the recall of selected views, and the other lies in the order of relevant views. Gemini-1.5-pro presents higher recall on Shift-H due to higher selection recall. Idefics3 maintains the same recall for all different settings, but achieves a higher accuracy on Shift-H. We hypothesis that the gain comes from the order of input views, where hard-level evaluation starts with less relevant views while appends more informative views at the end of input image sequence when all the views are selected. Please refer to Appendix~\ref{app:shift} for detailed analysis on shifting evaluation.

\subsection{Analysis of View Processing Strategies} 
We investigate these two aspects in Appendix~\ref{app:split} and Appendix~\ref{app:caption}, respectively. 
For the splitting settings, the adopted 4 sub-image setting provides fair and reliable evaluation results, which is not only effective and efficient, but also demonstrate a good balance between zooming and shifting evaluations. For the strategy of converting image into text, on the contrary, we observe significant drops of results on both zooming and shifting evaluations for most of investigated models. This suggests that the resizing issue in image concatenation strategy has only a minor impact on the performance. Please refer to Appendix~\ref{app:split} and Appendix~\ref{app:caption} for details.

\section{Conclusion}
This paper introduces ActiView, a novel benchmark designed to evaluate the active perception abilities of MLLMs. ActiView simulates real-world scenarios by imposing view constraints on images, requiring models to perform view shifting and/or zooming to gather necessary information for answering questions. Our results indicate that current MLLMs exhibit significantly lower active perception capabilities compared to humans, and that active perception abilities of models will be markedly enhanced by allowing inputs in multi-image interleaved structures. We also observed that models tend to perform better on our zooming evaluations compared to shifting evaluations. This suggests that the evaluated models lack the ability to combine their understandings of constrained perceptual fields to form a holistic perspective of the complete image or the full scene. We hope our benchmark will inspire further research in this critical area.


\section*{Limitations}
In this study, we utilize the form of VQA and intermediate reasoning behaviors to assess active perception abilities of models. While pipeline presents significant challenges for current multimodal language models (MLLMs), it does not encompass all aspects of active perception. For instance, it overlooks factors such as perspective distortion, multi-sensor integration, the incorporation of more dynamic or interactive environments, and real-time manipulation. With the development of reasoning ability of models such as OpenAI o1, o3 and DeepSeek-R1, we believe the active perception ability will also emerge or be improved. Moreover, techniques like tool learning and multi-agent collaboration could potentially enhance active perception performance based on existing MLLMs, making these areas worthy for future exploration and improvement. Additionally, active perception could also be used to improve other challenging topics such as hallucination and real-life application such as auto-driving. Also, Considering that these exceed the scope of a single conference paper, we do not include them in this paper, and we solely evaluates the inherent active perception capabilities of the MLLMs, aiming at an in-depth investigation of our primary focus. 
\bibliography{custom}

\clearpage
\appendix

\section{Data Details} \label{app:data}

\subsection{Image Collection}
To ensure the clearness of useful visual details, the collected original image should be of high resolution. In practice, we collected images of three resolution levels, including 1920 $\times$ 1040, 2250 $\times$ 1500, and 5184 $\times$ 3456, which are originated from VCR dataset~\citep{zellers2019recognition}, SA-1B dataset~\citep{kirillov2023segment}, and photos taken in daily life. At the beginning of the image collection process, 30 images are collected from photos taken from daily activities, which are then served as pilots and standards for manually expanding the data scale from SA-1B dataset~\citep{kirillov2023segment} and VCR dataset~\citep{zellers2019recognition}. These images should include rich and fine-grained visual details. 

\subsection{Rules for Annotation}

We provide a concise version of instruction used during annotation. For each of the images, annotators should follow the following instructions:
\begin{itemize}[left=0.3cm, itemsep=0.1pt, parsep=0.1pt]
    \item Questions: (1) Questions should be objective which have one and only one answer regarding the images. (2) The participation of multiple visual clues are preferred. They can be in the same or different regions of the image. 
    \item Options: (1) Options originated from the image itself are preferred. (2) The numeric options should be arrange randomly, neither descending nor ascending order. (3) Options cannot be opposite to each other, except for ``Yes'' or ``No''. (4) The number of options are not restricted to 4, you can provide as many options as long as they are reasonable and are closely related to the question and the image.
    \item Distraction: Annotator should provide distracting visual clues that could lead to wrong answer (if any).
    \item Clues: regions in the image that contribute to your annotated answer. 
\end{itemize}

\subsection{Detailed Category Description} \label{app:detail_cate}
When perceiving an image, humans intuitively focus on three principle aspects: the environment depicted in the image, the primary objects, and the event that these objects are engaged in. Correspondingly, we summarize the questions in our benchmark into three main categories, environment-centric (Type I), object-centric (Type II), and event-centric (Type III) categories, which are further divided into eight sub-classes according to the specific type of visual information and visual features used for answering the questions, as displayed in Figure~\ref{fig:example}.
For the environment-centric category, three sub-classes are developed:
\begin{itemize}[left=0.3cm, itemsep=0pt, parsep=0pt]
    \item \textbf{Geo-Localization} (\textit{Geo-Loc}) focuses on geographical features that are unique to a country or a city, and requires models to identify geographical locations depicted in target images. Typical questions are ``\textit{Where is this place located?}'',``\textit{In which country is the photo taken?}'', and etc. Images in this class usually contain unique landmarks such as the Eiffel Tower in Paris and the Atomium in Brussels.
    \item \textbf{Orientation} (\textit{Orient}) challenges models to exploit natural orientation information for answering the questions, such as the position of shadows, the position of the sun, and the directional information on street signs. Questions of this type include ``\textit{Is this a sunset or a sunrise?}'', ``\textit{Where is the sunlight coming from?}'' and etc.
    \item \textbf{Daily-location} (\textit{Daily-Loc}). To distinguish from Geo-Loc, this sub-class concentrates on locations in everyday life that could appear in most of the cities and are not unique to a certain city or country. Images in this sub-class usually depict scenes of museums, restaurants, shops, etc. The corresponding questions include ``\textit{Where is this picture most likely taken?}'', ``\textit{Is there a music school nearby?}'', and etc.
\end{itemize}

For the object-centric category, we expect models to exhibit abilities beyond simple grounding tasks that directly ask for the attributes or relations of objects. Questions for this category usually involve distracting information from images, and require models to precisely understand the intentions. Sub-classes are demonstrated as follows:
\begin{itemize}[left=0.3cm, itemsep=0pt, parsep=0pt]
    \item \textbf{Object-attribute} (\textit{Obj-Attr}) addresses objects attributes while distracting information, that potentially lead to incorrect answers, appears in the images. Shown by the Obj-Attr case in Figure~\ref{fig:example}, the highest price, 69 per kilogram, corresponds to papaya rather than watermelon.
    \item \textbf{Object-relation} (\textit{Obj-Rel}) concentrates on the spatial relationships among multiple objects, while the questions do not directly ask for the spatial relationships. It requires models to reason for the correct answer via spatial information. Figure~\ref{fig:example} displays an Obj-Rel case in which models should be aware of the relative positions of the feet of the person to the water. 
    \item \textbf{Counting} (\textit{Count-Dis}). Although it focuses on the number of objects, different from the counting tasks in other benchmarks~\citep{fu2023mme, yu2023mm}, there are similar but distracting information about the targets in our images. These distracting objects easily confuse models and challenge the abilities to understand and strictly follow instructions. As the Count-Dis case in Figure~\ref{fig:example}, the jerky on the table are distracting to the answer of question ``\textit{How many pieces of jerky are hanging on the wall?}''.
\end{itemize}

The event-centric category focuses on the interactions of humans and items, such as movements, actions and activities. This category is divided according to the number of objects involved in the target event as following:
\begin{itemize}[left=0.3cm, itemsep=0pt, parsep=0pt]
    \item \textbf{Event-single} (\textit{Event-S}). There is only one item or person involved in the target event. For example, the image for Event-S in Figure~\ref{fig:example} shows one person driving without other people presenting in the image. 
    \item \textbf{Event-multi} (\textit{Event-M}). Different from Event-S, events of this type happen among multiple items or people. In the Event-M case in Figure~\ref{fig:example}, the ``\textit{woman in blue}'' is engaged in a photo shooting activity in which she is posing and another person is taking photo for her. It requires models to distinguish the event or events that each entities are engaged in.
\end{itemize}

\section{Human Evaluation}\label{app:human}

\begin{table}
\centering\small
\begin{tabular}{@{\hspace{0.1cm}}c@{\hspace{0.1cm}}|@{\hspace{0.1cm}}c@{\hspace{0.1cm}}|cc|@{\hspace{0.1cm}}c@{\hspace{0.1cm}}}
\toprule
Annotator & Background & ACC & ACC* & Consis. \\ \midrule
User1 & CS & 73.33 & 85.00 & 100.00 \\ 
User2 & Med & 71.67 & 75.00 & 93.33 \\ 
User3 & Telecom & 85.00 & 90.00 & 100.00 \\ 
User4 & CS & 81.67 & 88.33 & 83.33 \\ 
User5 & CS & 76.67 & 85.00 & 95.00 \\
User6 & Art & 70.00 & 78.83 & 83.33 \\
\midrule
\multicolumn{2}{c@{\hspace{0.1cm}}|}{Average}& 77.53 & 84.67 & 94.20 \\
\bottomrule
\end{tabular}
\caption{Human level performance and question consistency. Consis.: human-annotated consistency of question-image-option-groundtruth. ACC: accuracy of answering the questions without assistance (\textit{i.e.}, the accuracy for ``Human'' evaluation). ACC*: accuracy of answering the questions with the help of Internet (\textit{i.e.}, the accuracy for ``Human*'' evaluation).} 
\label{tab:human_detail}
\end{table}

We sampled 60 questions for human-level test, and recruit 6 testees, who did not participate in image collection and question annotation, to evaluate the human level performance of our benchmark. These testees are from diverse backgrounds, including computer science (CS), telecommunication (Telecom), Medicine (Med), and Art. 

\begin{table*}[t]
    \centering\small
    \begin{tabular}{l|cccccc}
    \toprule
    Model&	Claude&	GPT-4o	&Qwen2-VL	&MiniCPM-V 2.6	&Idefics3	&Brote-IM-XL \\
    \midrule
    ACC	& 2.14	&2.45&	23.38&	26.77&	44.92	&40.00 \\
    ACC(guess)&	26.07&	37.73	&42.77&	41.54	&47.38	&40.00 \\
    \bottomrule
    \end{tabular}
    \caption{Results of text-only evaluation. ACC: answer with commonsense only without random guessing. ACC(guess): guess the answer according to commonsense.} \label{tab:textonly}
\end{table*}

For a fair comparison with the MLLMs, we employ two settings, including a ``Human'' evaluation that asks testees to answer questions all by themselves, and a ``Human*'' evaluation that allows testees to use the Internet and LLMs for the required knowledge, because these testees may not be exposed to knowledge that never appear in their everyday life, which MLLMs have already seen in the training data. Note that in the ``Human*'' evaluation, directly search for the answer to the questions are forbidden. Referring to the question in Figure~\ref{fig:pipeline} as an example, testees may search for ``\textit{what does the national flag of UK/France/Spain/Hungary look like?}'', which may provide extra knowledge that helps them to answer the original question.
Manual evaluation achieves an average accuracy of 84.67\%, which is more than doubled of the random choice result (33.95\%), while some models present only slightly higher accuracies compared to the random result. These indicate the potential for models to get improved.

We ask testees to vote for the consistency of the annotated question-image-option-groundtruth quadruple for the investigation of the reliability of our benchmark. The consistency score represents if the testee agree with these quadruples and find the groundtruth answers and the provided options are practical and reasonable. Our benchmark is reliable indicated by a consistency of 94.2\%.

\subsection{Analysis of Human Performances}
We assess the difficulty and reliability of our benchmark upon human performances in Table~\ref{tab:human_detail}. 
We employ two settings for human evaluation, where ``Human'' asks annotators to answer questions only by themselves, and ``Human*'' allows annotators to use the Internet and LLMs for extra knowledge that could help answer the questions. This ``Human*'' evaluation aims at fair comparison as humans may not be exposed to knowledge that never appear in their everyday life, while most of MLLMs should be aware of these knowledge from the training data. 
Human presents an average accuracy of 77.53\%, suggesting that our benchmark is challenging even for human. For a fair comparison with large models, ``Human*'' achieves an average of 84.67\% by allowing searching for world knowledge from the Internet or using LLMs. 
Human performances (84.67\%) are more than doubled of the random result (33.95\%), while some models present only slightly higher accuracies compared to the random result. These indicate the potential for models to get improved.

\begin{table*}[t!]
\centering\small
\begin{tabular}{l|c|c}
\toprule
Models & LLM Backbone & Vision Encoder \\
\midrule \noalign{\vskip -3pt}
\multicolumn{3}{c}{\cellcolor[HTML]{EFEFEF} \scriptsize\textit{APIs}}    \\ \noalign{\vskip 1pt}
GPT-4o~\citep{gpt-4o} & \multicolumn{2}{l}{gpt-4o} \\
Gemini-1.5-pro~\citep{reid2024gemini}& \multicolumn{2}{l}{gemini-1.5-pro} \\ 
Claude 3.5 Sonnet ~\citep{TheC3}&  \multicolumn{2}{l}{claude-3-5-sonnet-20240620} \\
\midrule\noalign{\vskip -3pt}
\multicolumn{3}{c}{\cellcolor[HTML]{EFEFEF} \scriptsize\textit{Open-Source Models}}    \\ \noalign{\vskip 1pt}
GLM-4V-9B ~\citep{du2022glm}& GLM-4-9B & CLIP \\
SEAL~\citep{wu2023textit} & Vicuna-7B & CLIP ViT-L/14\\
InternVL-Vicuna-7B~\citep{chen2023internvl} & Vicuna-7B&InternViT \\
InternVL-Vicuna-13B~\citep{chen2023internvl} & Vicuna-13B & InternViT \\
InternVL-Vicuna-13B-448px~\citep{chen2023internvl}& Vicuna-13B &InternViT‑300M‑448px  \\
InternVL2-8B~\citep{chen2024far} &internlm2\_5‑7b‑chat &InternViT‑300M‑448px\\
MiniCPM-Llama3-V-2.5~\citep{yao2024minicpm} & Llama-3-8B & SigLip-400M   \\
MiniCPM-V 2.6 ~\citep{yao2024minicpm}& Qwen2-7B & SigLip-400M   \\
LLaVA-1.6-13B~\citep{liu2024llavanext} & Vicuna-13B & CLIP-ViT-L/14  \\
LLaVA-1.6-7B~\citep{liu2024llavanext} & Vicuna-7B & CLIP-ViT-L/14 \\
LLaVA-OneVision-7B ~\citep{li2024llavaov}& Qwen2-7B&SO400M\\
Phi-3.5-Vision~\cite{abdin2024phi3} & Phi-3.5 & CLIP-ViT-L-16-336 \\
mPLUG-Owl2-7B ~\citep{ye2024mplug}& Llama-2-7B & CLIP ViT-L/14 \\
mPLUG-Owl3-7B ~\citep{ye2024mplug3}& Qwen2-7B & Siglip-400m \\
Qwen2-VL-8B~\citep{wang2024qwen2vl} & Qwen2-7B & OpenCLIP-ViT-bigG \\
Qwen2.5-VL-3B~\citep{Qwen2.5VL} & Qwen2.5 & trained from scratch \\
Qwen2.5-VL-7B~\citep{Qwen2.5VL} & Qwen2.5 & trained from scratch \\
Deepseek-VL-7B~\citep{lu2024deepseekvl} & Deepseek & Siglip-large-patch16-384 \\
Deepseek-VL2~\citep{wu2024deepseekvl2mixtureofexpertsvisionlanguagemodels} & Deepseek2 & Siglip-400m \\
Mantis~\citep{jiang2024mantis} & LLaMA-3  & Siglip-400m \\
Idefics2-8B~\citep{laurenccon2024matters} &Mistral-7B & Siglip-400m \\
Idefics2-8B-base~\citep{laurenccon2024matters} &Mistral-7B & Siglip-400m \\
Idefics3-8B-Llama3\citep{laurençon2024idefics3}&Mistral-7B & Siglip-400m \\
MMICL-XXL~\citep{zhao2023mmicl} & FlanT5-XXL-11B& EVA-G \\
Brote-IM-XXL~\citep{wang2024browse} & FlanT5-XXL-11B & EVA-G\\
MMICL-XL~\citep{zhao2023mmicl} & FlanT5-XL-3B & EVA-G\\
Brote-IM-XL~\citep{wang2024browse} & FlanT5-XL-3B & EVA-G\\
Mini-Gemini-7B-HD~\citep{li2024mini}&LLaMA-3&CLIP-L \\
Mini-Gemini-7B~\citep{li2024mini} &LLaMA-3&CLIP-L\\
\bottomrule
\end{tabular}
\caption{The versions of LLM backbone and vision encoder of our evaluated models. For proprietary models, we provide the API version we used.} 
\label{tab:models}
\end{table*}

\section{Text-only Evaluation}\label{app:textonly}
We provide a text-only evaluation to measure the amount of commonsense answers with providing images in our benchmark. We conducted two experiments: 
\begin{itemize}[left=0.3cm, itemsep=0.1pt, parsep=0.1pt]
    \item Commonsense-only evaluation. This evaluation aims at measuring the amount of questions that can be answered only via commonsense knowledge without searching for visual clues in the image. The template is as follows: ``Please answer questions based on you commonsense knowledge. If you are not able to answer the question based soly on the commonsense knowledge you've acquired, please \textbf{response with `None'}. {Question} {Options} Your answer:''
    \item Commonsense and data bias evaluation. Considering that current models are trained with a large amount of data and various tasks, they could potentially memories the most frequent answers given a image-question pair. We implement another template to evaluate the amount of data that can be correctly guessed without corresponding context. The template is as follows: ``Please answer questions based on you commonsense knowledge. If you are not able to answer, please \textbf{select a most probable one}. {Question} {Options} Your answer:''
\end{itemize}

Results for these text-only evaluations are listed in Table~\ref{tab:textonly}. This table indicates that questions in our benchmark cannot be simply answered via commonsense, where two powerful models GPT-4o and Claude achieves only 2.45\% and 2.14\% for commonsense-only evaluation. The row of ACC(guess) presents results of generating the most probable answers, reflecting the bias obtained from the training corpus.
The differences between these two type of evaluation are caused by the ability of instruction-following. We found that Idefics3 and Brote-IM-XL present weaker instruction-following ability compared to other models in this table, that they still exhibit a behavior of guessing when commonsense cannot be used to answer the questions.

Overall, our benchmark requires elaborate observation of the given images and comprehensive understanding of image-question pairs, which cannot be solved simply by commonsense.

\section{Models}\label{app:model}

We investigate both proprietary and open-source models. The proprietary models include widely discussed 
GPT-4o~\citep{gpt-4o}, Gemini-1.5-pro~\citep{reid2024gemini}, and Claude 3.5 Sonnet~\citep{TheC3}. For open-source models, we carefully select recent and commonly used models of different structures and of difference scales, such as model families of MiniCPM-V~\cite{yao2024minicpm}, LLaVA~\citep{liu2023improved, liu2023visual}, mPLUG-Owl~\citep{ye2024mplug, ye2024mplug3}, Idefics~\citep{laurenccon2024matters, laurençon2024idefics3}, and etc. 
Since the awareness of fine-grained details and instruction-aware visual features are significant indicators during evaluation, we also include models specifically optimised on these aspects, such as SEAL~\citep{wu2023textit} for fine-grained details understanding, and Brote~\citep{wang2024browse} which is trained from InstructBLIP~\citep{InstructBLIP} for instruction-aware and multi-image comprehension. 
Details of these models are listed in Table~\ref{tab:models}. Considering models of different scales, we include a total of 27 models. These models are divided into two types, \textbf{single-image models} that accepting only one image per input, such as LLaVA-1.6~\citep{liu2023improved} and MiniCPM-Llama3-V-2.5~\citep{yao2024minicpm}; and \textbf{multi-image models} that allow more than one images to appear in the same input, such as Brote and Idefics. 
We describe the approaches for integrating multiple views into the input for the two types of models in Appendix~\ref{app:process} and Appendix~\ref{app:caption}.

\section{Evaluation Pipelines} \label{sec:pipeline}
This section will discuss motivations and settings of each pipelines in detail.

\vspace{-6pt}
\paragraph{Zooming pipeline.} It focuses on one of the fundamental factors, zooming, and evaluates the ability to locate and determine fine-grained information necessary to answer questions. As illustrated in Figure~\ref{fig:pipeline} (a), this pipeline contains two stages, the view selection and the question answering stages. To simulate the zooming operation, models are required to first select sub-views to be zoomed given the initial view, then answer questions based on these zoomed views. The initial view used in this pipeline is the full image with size $w \times h$. Each of the selected sub-views will be resized to size $w \times h$, the same as the initial view. In Figure~\ref{fig:pipeline} (a), the zoomed right-upper view is resized as a $w \times h$ image, and so does the zoomed left-lower view. Afterwards, models answer the question given the two zoomed views. 
Please refer to Appendix~\ref{app:template_zoom} for prompt templates.

\vspace{-6pt}
\paragraph{Shifting pipeline.} It addresses the other fundamental factor, shifting, and emphasizes the ability to navigate perceptual fields incrementally, mimicking real-world scenarios where full context is unavailable. It evaluates the ability to shift perceptual fields for missing information and to deduce the answer given perceived perceptual fields following templates in Appendix~\ref{app:template_shift}. 
This is also a two-stage pipeline as in Figure~\ref{fig:pipeline} (b). To simulate the movement of human eyes, models are presented with an initial view, size $w \times h$, which is a cropped field from the original image, and are asked to determine if the current views are sufficient for answering. Upon receiving positive responses, models are prompted to produce answer given the current view or views. If the model requires more views to infer the answer, an adjacent view will be given until the model can answer the question. 
For this pipeline, we further assign different difficulties according to human-annotated visual clues contained in the initial views as follows:
\begin{itemize}[left=0.3cm, itemsep=1pt, parsep=1pt]
    \item Shift-R: randomly selected initial views.
    \item Shift-E: easy-level evaluation, where initial views contain at least one entire visual clue for answering the question.
    \item Shift-M: medium-level evaluation, where initial views contain only partial visual clues for answering the question. 
    \item Shift-H: hard-level evaluation, where no visual clues appear in the initial views.
\end{itemize}

\paragraph{Mixed pipeline.} While the above pipelines permit either zooming or shifting individually, we also implement an automated mixed setting that does not specify the type of active perception ability required. As illustrated in Figure~\ref{fig:pipeline} (c), models must independently decide whether to zoom and/or shift to different perceptual fields. Unlike the zooming pipeline, where the model answers questions based on all selected views, in the mixed pipeline, a view would be discarded after selection if the model recognizes it as irrelevant to the question. Compared to the shifting pipeline, the mixed pipeline also provides access to the full image view in addition to cropped sub-views. Appendix~\ref{app:template_mix} records the employed prompt templates.
This pipeline requires models to account for all the sub-views and the full image for unbiased operation determination and view selection. Otherwise, it is at risk of reverting to zooming or shifting evaluation without sufficient and unconverted visual information. Therefore, the mixed pipeline emphasizes the autonomy of models and is only applied to multi-image models.

\section{Discussion on Image Splitting and Processing Strategies} \label{app:image_process}
\subsection{Image Splitting Settings}\label{app:split}
In our final pipelines, the original images are equally split into 4 views. We also conduct experiments of splitting into more views and report the results in Table~\ref{tab:split}. We found that the 4 sub-image setting is able to derive fair and reliable evaluation results, which is not only effective but also efficient. More splits require additional inference time and resources (e.g., the context length, GPU memory, etc.), but they only yield similar trends and conclusions compared to 4 sub-image setting. 

\begin{table}[t]
    \centering\small
    \begin{tabular}{@{\hspace{0.1cm}}l@{\hspace{0.1cm}}|@{\hspace{0.1cm}}c@{\hspace{0.1cm}}|@{\hspace{0.1cm}}c@{\hspace{0.15cm}}c@{\hspace{0.1cm}}|@{\hspace{0.1cm}}c@{\hspace{0.1cm}}}
        \toprule
        Model & Splits & Zooming & Shift-R & AVG \\
        \midrule
        LLaVA-1.6 7B & 4 & 68.92 & 51.69 & 60.31 \\
        LLaVA-1.6 7B & 6 & 73.23 & 53.85 & 63.54\\
        LLaVA-1.6 7B & 8 & 72.92 & 48.61 & 60.77 \\
        LLaVA-1.6 7B & 9 & 66.46 & 46.16 & 56.31 \\
        LLaVA-1.6 7B & 16 & 69.23 & 46.15 & 57.69 \\
        \midrule
        LLaVA-1.6 13B & 4 & 65.23 & 53.85 & 59.54 \\
        LLaVA-1.6 13B & 6 & 71.69 & 46.46 & 59.07\\
        LLaVA-1.6 13B & 8 & 71.84 & 44.00 & 57.92 \\
        LLaVA-1.6 13B & 9 & 72.00 & 43.69 & 57.84 \\
        LLaVA-1.6 13B & 16 & 73.31 & 43.23 & 58.27 \\
        \bottomrule
    \end{tabular}
    \caption{Experimental results of different splits.}
    \label{tab:split}
\end{table}

Additionally, there are two issues with more splits. First, it is challenging for the ability to process multiple images and understand their relationships. As shown in the table above, when increasing the number of splits, LLaVA-1.6-7b degrades from 60.31 to 57.69 (-2.62) on average, and LLaVA-1.6-13b decreases 1.27 on average. Although increasing the splits would increase the performance of zooming evaluation, the performance of shifting is remarkably decreased. As we focus on active perception concerning both zooming and shifting, a split of 4 would present a decent balance. Second, the necessary information would be more likely to be split into different tiles, causing information loss.

\begin{table*}[t]
    \centering\small
    \begin{tabular}{l|c|ll}
        \toprule
        Model & Visual Info. Type & Zooming & Shift-R  \\
        \midrule
        LLaVA-1.6 7B & Image concatenation & 68.92 & 51.69  \\
         & Textual descriptions & 60.31 \scriptsize{\textcolor{red}{-8.61}} & 53.83 \scriptsize{+2.14} \\
         \midrule
        LLaVA-1.6 13B & Image concatenation  & 65.23 & 45.85 \\
         & Textual descriptions & 60.00 \scriptsize{\textcolor{red}{-5.23}} & 43.69 \scriptsize{\textcolor{red}{-2.16}} \\
         \midrule
        mPLUG-Owl2 7B & Image concatenation & 55.38 & 47.38  \\
         & Textual descriptions & 62.77 \scriptsize{+7.39} & 54.15 \scriptsize{+6.77}\\
        \midrule
        MiniCPM-Llama3-V-2.5 & Image concatenation & 61.25 & 60.92 \\
         & Textual descriptions & 61.25 \scriptsize{\textcolor{red}{-0}} & 60.31 \scriptsize{\textcolor{red}{-0.61}}\\
        \bottomrule
    \end{tabular}
    \caption{Experimental results providing single-image models with captions as compensation for the invisibility of previous images.}\label{tab:desc}
\end{table*}

\subsection{Processing of Views} \label{app:process}
The question answering stage of the zooming, shifting, and mixed pipelines, as well as the missing view examination stage of the shifting pipeline, require multi-image inputs if multiple views are selected. In this paper, we primarily focus on the interleaved multi-image setting, since it is more practical and natural compared to the single-image setting. Multi-image models can naturally read and understand multiple views at one time (in the form of different images) during evaluating, and we directly format the images and text in an interleaved format. However, we also propose methods for evaluating powerful single-image models. For these models, we employ two strategies to enable simultaneous understanding of different views. One is to concatenate the required views into a single flattened image, and the other preserves merely the current view as an image while converting the remainings into textual descriptions. The following subsection discusses this in detail.

\subsection{Strategies of Processing Multiple Images for Single-image Models}
\label{app:caption}

For all the pipelines, multiple views might be selected depending on the response of models, which can be naturally handled by multi-image models. However, for models that only accepts single image per input, we apply different image processing approaches for zooming and shifting pipelines. For the shifting pipeline, we proposed to concatenate the selected views or convert them into textual descriptions to fit the information of multiple images into a single input. The concatenation refer to stitch the images selected views together from left to right to form a single image as the input for the model. This is applicable for both missing view examination stage and question answering stage. For the question answering stage in zooming pipeline, if multiple views were selected in the first stage of our pipelines, we will use the each selected view to ask questions sequentially. After obtaining answers, if the model answers correctly based on any of the views, we consider it a complete and successful view selection. 

In addition to the directly processing of image, we also propose methods to deliver visual information by converting images into textual descriptions. This enables single-image models to ``see'' multiple images in the form of text inputs. This method can be applied to both shifting and zooming settings. When multiple views are required, we preserve merely the current view in the form of image, while converting the remainings into textual descriptions via the prompt ``Please describe the image:''. Results of typical single-image models, LLaVA-1.6, mPLUG-Owl2 and MiniCPM-Llama3-V-2.5 are shown in Table~\ref{tab:desc}.

For the strategy of converting image into text, it is supposed to be a compensation for the image concatenation strategy to avoid images being resized. On the contrary, we observe significant drops of results on both zooming and shifting evaluations for most of the investigated models, indicating that the resizing issue of image concatenation strategy has minor influence on the performance. Moreover, the operation to converting images into textual descriptions introduces the influence of other abilities that interferes the evaluation of active perception abilities.

\section{Measurements of View Selection}\label{app:view_measure}
We follow the recall, precision and F1 metrics to evaluate the performance of the view selection for zooming setting and the missing view examination for shifting settings. We denote the selected views containing human-annotated clues as $TP_{op}$, where $op$ refers to either ``zoom'', ``shift'' or ``mix''. $FN_{op}$ refers to views that contain human-annotated clues but are not selected for answering questions, and $FP_{op}$ refers to views selected but do not contain human-annotated clues. 
Finally, the precision, $P_{\text{select}}$, is calculated as follows:
\begin{equation}
\label{eq:pre}
P_{\text{select}} = \frac{TP_{op}}{TP_{op} + FP_{op}}, {\small \quad op \in \{\text{zoom}, \text{shift}, \text{mix}\}}.
\end{equation}
It measures the proportion of selected views that are actually relevant. It reflects the model ability to avoid unnecessary or irrelevant views. A higher precision indicates that the model is more efficient in identifying only the information necessary for answering the question.
The recall, $R_{\text{select}}$, is calculated as follows:
\begin{equation}
\label{eq:recall}
R_{\text{select}} = \frac{TP_{op}}{TP_{op} + FN_{op}}, {\small \quad op \in \{\text{zoom}, \text{shift}, \text{mix}\}}.
\end{equation}
This recall score measures the proportion of views correctly identified by the model out of all views containing human-annotated clues. It reflects the model ability to capture the required information. A higher recall indicates that the model is less likely to miss important views during the zooming process. Accordingly, F1 score of view selection, $F_1$ is computed as:
\begin{equation}
\label{eq:f1}
F_1 = \frac{2 \cdot P_{\text{select}} \cdot R_{\text{select}}}{P_{\text{select}} + R_{\text{select}}}
\end{equation}

\begin{table*}[h!]
\centering\scriptsize
\begin{tabular}{l@{\hspace{0.1cm}}|@{\hspace{0.1cm}}c@{\hspace{0.1cm}}|@{\hspace{0.1cm}}c@{\hspace{0.1cm}}|@{\hspace{0.1cm}}c@{\hspace{0.1cm}}|c@{\hspace{0.2cm}}c@{\hspace{0.1cm}}c@{\hspace{0.1cm}}c@{\hspace{0.1cm}}|@{\hspace{0.1cm}}c@{\hspace{0.1cm}}}
\toprule
\multirow{2}{*}[-1ex]{Models} & \multicolumn{2}{@{\hspace{0.1cm}}c@{\hspace{0.1cm}}|@{\hspace{0.1cm}}}{Zooming} & \multicolumn{5}{c@{\hspace{0.1cm}}|@{\hspace{0.1cm}}}{Shifting} & \multirow{2}{*}[-1ex]{\begin{tabular}[c]{@{}c@{}}Models\\AVG\end{tabular}} \\ \cmidrule(lr){2-3} \cmidrule(lr){4-8} 
& Full image & Zooming & Single View & Shift-R & Shift-E & Shift-M & Shift-H & \\
\midrule\noalign{\vskip -3pt}
\multicolumn{9}{c}{\cellcolor[HTML]{EFEFEF} \scriptsize\textit{proprietary models}}    \\ \noalign{\vskip 1pt}
Gemini-1.5-pro & \textbf{73.85} & \textbf{72.31} & 58.15 & \textbf{67.08 }& \textbf{67.38} & \textbf{65.54} & \textbf{67.69} & \textbf{68.00} \\
GPT-4o & 67.38 & 68.62 & \textbf{61.23} & \textbf{67.08} & 66.77 & 65.23 & 64.31 & 66.40  \\
Claude 3.5 Sonnet & 72.92 & 71.69 & 54.46 & 65.23 & 66.15 & 60.31 & 61.85 & 65.05    \\
\noalign{\vskip -2pt}
\midrule\noalign{\vskip -3pt}
\multicolumn{9}{c}{\cellcolor[HTML]{EFEFEF} \scriptsize\textit{Open-source models for multiple images as input}}    \\ \noalign{\vskip 1pt}
Qwen2.5-VL-7B & \colorbox{gray!10}{\textcolor{gray}{67.08}} & \colorbox{gray!10}{68.92} & 47.69 & \colorbox{gray!10}{68.62} & 67.08 & 67.38 & 68.00 & 68.00 \\
Qwen2.5-VL-3B & \textcolor{gray}{65.85} & \colorbox{gray!10}{66.15} & 55.08 & 65.32 & 65.85 & 65.54 & 65.32 & 65.64 \\
DeepSeek-VL2 & 70.46 & 65.85 & 58.15 & 65.54 & 65.23 & 64.31 & 64.62 & 65.11 \\
Qwen2-VL & 63.08 & 64.62 & 54.46 & 61.23 & \colorbox{gray!10}{62.77} & \colorbox{gray!10}{64.31} & \colorbox{gray!10}{61.85} & \colorbox{gray!10}{62.96} \\
Idefics3-8B-Llama3 & 59.08 & 58.15 & 53.23 & \colorbox{gray!10}{61.85} & 59.38 & 59.69 & 60.31 & 59.88 \\
MiniCPM-V 2.6 & 64.62 & 61.85 & 54.46 & 54.77 & 61.23 & 58.15 & 55.69 & 58.34 \\
mPLUG-Owl3 & 62.46 & 60.92 & 54.15 & 51.69 & 56.31 & 55.69 & 53.54 & 55.63 \\
LLaVA-OneVision & \colorbox{gray!10}{64.92} & \colorbox{gray!10}{65.23} & \colorbox{gray!10}{56.92} & 53.54 & 57.23 & 52.31 & 48.62 & 55.39 \\
InternVL2-8B & 58.15 & 56.00 & 45.85 & 54.77 & 59.70 & 53.23 & 52.00 & 55.14  \\
Mantis & 59.08 & 60.62 & 52.92 & 52.92 & 55.38 & 52.92 & 52.31 & 54.83 \\
Idefics2-8B & 61.85 & 61.85 & 55.69 & 53.23 & 56.92 & 51.69 & 49.23 & 54.58 \\	
Brote-IM-XL-3B & 54.77 & 54.46 & 55.69 & 51.38 & 51.08 & 52.62 & 47.69 & 51.45 \\
Phi-3.5-Vision & 55.08 & 56.62 & 48.92 & 50.46 & 54.15 & 50.15 & 45.54 & 51.38 \\
DeepSeek-VL-7B & 53.23 & 53.23 & 49.85 & 50.15 & 49.85 & 51.69 & 51.69 & 51.32 \\
Idefics2-8B-base & 52.62 & 48.62 & 47.69 & 49.54 & 50.77 & 47.69 & 47.69 & 48.86 \\
Brote-IM-XXL-11B & 53.85 & 54.77 & 49.23 & 49.85 & 50.77 & 44.92 & 43.69 & 48.80 \\
MMICL-XXL-11B & 51.69 & 49.54 & 50.15 & 49.85 & 49.85 & 46.77 & 45.54 & 48.31 \\
MMICL-XL-3B  & 49.85 & 49.85 & 44.31 & 44.92 & 48.92 & 45.85 & 44.31 & 46.77 \\ \noalign{\vskip -2pt}
\midrule\noalign{\vskip -3pt}
\multicolumn{9}{c}{\cellcolor[HTML]{EFEFEF} \scriptsize\textit{Open-source models for single image as input}}    \\ \noalign{\vskip 1pt}
MiniCPM-Llama3-V-2.5 & 63.87 & 61.25 & \colorbox{gray!10}{54.47} & \colorbox{gray!10}{60.92} & 60.31 & \colorbox{gray!10}{59.38} & \colorbox{gray!10}{58.46} & \colorbox{gray!10}{60.06} \\ 
GLM-4V-9B & \colorbox{gray!10}{67.08} & 56.92 & 53.85 & 56.92 & \colorbox{gray!10}{60.62} & 56.00 & 52.92 & 56.68 \\
InternVL-Vicuna-13B & 56.92 & 62.77 & 52.31 & 53.85 & 52.92 & 52.92 & 51.08 & 54.71 \\
LLaVA-1.6 7B & 55.08 & \colorbox{gray!10}{68.92} & 50.15 & 51.69 & 52.31 & 49.23 & 48.00 & 54.03 \\
InternVL-Vicuna-7B & 55.38 & 65.23 & 51.70 & 52.92 & 51.38 & 50.77 & 48.62 & 53.78 \\
LLaVA-1.6 13B & 56.92 & 65.23 & 52.31 & 45.85 & 55.08 & 52.62 & 48.92 & 53.54 \\
InternVL-Vicuna-13B-448px & 50.46 & 57.85 & 45.54 & 48.31 & 48.31 & 48.92 & 48.92 & 50.46 \\
mPLUG-Owl2-7B & 55.08 & 55.38 & 52.00 & 47.38 & 46.46 & 46.46 & 46.15 & 48.37 \\
Mini-Gemini-7B-HD & 55.69 & 34.77 & 51.70 & 48.62 & 48.00 & 47.69 & 50.15 & 45.85 \\
SEAL & 48.31 & 54.77 & 42.77 & 42.15 & 42.77 & 40.02 & 40.62 & 44.07  \\
Mini-Gemini-7B & 47.08 & 17.85 & 47.38 & 39.38 & 38.15 & 38.15 & 36.00 & 33.91   \\ \noalign{\vskip -2pt}
\bottomrule
\end{tabular}
\caption{The evaluation of active perception abilities on our benchmark, including zooming (for limited resolution scenarios), and shifting (for scenarios of limiting the field of views). ``Model AVG'': average scores of column ``Zooming'', ``Shift-R'', ``Shift-E'', ``Shift-M'', and ``Shift-H''. The best scores of each column are \textbf{bolded} and the best scores in each model types are \colorbox{gray!10}{highlighted}.} 
\label{tab:full_rst}
\end{table*}

\section{Experimental Results}\label{app:more_rst}
We reported the full results of 27 models in Table~\ref{tab:full_rst}. This table preserves the conclusions as discussed in by Table~\ref{tab:rst_}. The detailed results of each categories are listed in Table~\ref{tab:zoom}, Table~\ref{tab:shiftR}, Table~\ref{tab:shift_easy}, Table~\ref{tab:shift_medium}, and Table~\ref{tab:shift_hard}, for zooming, Shift-R, Shift-E, Shift-M, and Shift-H, respectively.

\begin{table*}[h!]
\centering\scriptsize
\begin{tabular}{@{\hspace{0.05cm}}l@{\hspace{0.05cm}}|c@{\hspace{0.05cm}}c@{\hspace{0.05cm}}c@{\hspace{0.05cm}}c|c@{\hspace{0.05cm}}c@{\hspace{0.05cm}}c@{\hspace{0.1cm}}c|c@{\hspace{0.05cm}}c@{\hspace{0.1cm}}c@{\hspace{0.05cm}}}
\toprule
\multirow{2}{*}[-1ex]{Models} & \multicolumn{3}{c}{Type I} & \multirow{2}{*}[-1ex]{AVG} & \multicolumn{3}{c}{Type II} & \multirow{2}{*}[-1ex]{AVG} & \multicolumn{2}{c}{Type III} & \multirow{2}{*}[-1ex]{AVG} \\ \cmidrule(lr){2-4} \cmidrule(lr){6-8} \cmidrule(lr){10-11} 
& Geo-Loc & Orient & Daily-Loc &  & Obj-Attr & Obj-Rel & Count-Dis & & Event-M & Event-S & \\ \midrule\noalign{\vskip -3pt}
\multicolumn{12}{c}{\cellcolor[HTML]{EFEFEF} \scriptsize\textit{proprietary models}}    \\ \noalign{\vskip 1pt}
Gemini-1.5-pro & 91.89 & 60.00 & \textbf{92.68} & \textbf{83.33} & 80.00 & 65.22 & \textbf{51.06} & 65.73 & \textbf{85.29} & 57.50 & 70.27\\
GPT-4o & 94.59 & 63.33 & 85.37 & 82.41 & 68.00 & 54.35 & 46.81 & 56.64 & 76.47 & 65.00 & 70.27 \\
Claude 3.5 Sonnet & \textbf{97.30} & 50.00 & 87.80 & 80.56 & 72.00 & 67.39 & 42.55 & 60.84 & 82.35 & 75.00 & \textbf{78.38} \\
\midrule\noalign{\vskip -3pt}
\multicolumn{12}{c}{\cellcolor[HTML]{EFEFEF} \scriptsize\textit{Open-source models for multiple images as input}}    \\ \noalign{\vskip 1pt}
Qwen2-VL  & \textbf{97.30} & 50.00 & 80.49 & 77.78 & 68.00 & 65.22 & 40.43 & 58.04 & 58.82 & 57.50 & 58.11\\
Idefics3-8B-Llama3  & 89.19 & 56.67 & 73.17 & 74.07 & 60.00 & 54.35 & 29.79 & 48.25 & 58.82 & 50.00 & 54.05 \\
MiniCPM-V 2.6  & 86.49 & 46.67 & 80.49 & 73.15 & 54.00 & 56.52 & 31.91 & 47.55 & 61.76 & 42.50 & 51.35 \\
mPLUG-Owl3  & 89.19 & 53.33 & 80.49 & 75.93 & 64.00 & 60.87 & 36.17 & 53.85 & 58.82 & 47.50 & 52.70\\
LLaVA-OneVision  & 91.89 & 46.67 & 87.80 & 77.78 & 74.00 & 58.70 & 42.55 & 58.74 & 61.76 & 57.50 & 59.46 \\
InternVL2-8B & 75.68 & 56.67 & 70.73 & 68.52 & 60.00 & 47.83 & 25.53 & 44.76 & 61.76 & 57.50 & 59.46 \\
Mantis  & 89.19 & 41.38 & 80.00 & 72.64 & 72.00 & 54.35 & 44.68 & 57.34 & 54.55 & 51.28 & 52.78 \\
Idefics2-8B & 89.19 & 63.33 & 85.37 & 80.56 & 72.00 & 50.00 & 40.43 & 54.55 & 55.88 & 45.00 & 50.00 \\	
Brote-IM-XL-3B & 86.49 & 40.00 & 73.17 & 68.52 & 60.00 & 43.48 & 40.43 & 48.25 & 44.12 & 47.50 & 45.95\\
Idefics2-8B-base & 89.19 & 56.67 & 78.05 & 75.93 & 42.00 & 39.13 & 29.79 & 37.06 & 23.53 & 35.00 & 29.73 \\
Brote-IM-XXL-11B  & 86.49 & 33.33 & 80.49 & 69.44 & 58.00 & 43.48 & 34.04 & 45.45 & 58.82 & 45.00 & 51.35 \\
MMICL-XXL-11B  & 67.57 & 53.33 & 65.85 & 62.96 & 52.00 & 36.96 & 34.04 & 41.26 & 58.82 & 35.00 & 45.95 \\
MMICL-XL-3B  & 70.27 & 43.33 & 68.29 & 62.04 & 58.00 & 34.78 & 36.17 & 43.36 & 38.24 & 50.00 & 44.59 \\
\midrule\noalign{\vskip -3pt}
\multicolumn{12}{c}{\cellcolor[HTML]{EFEFEF} \scriptsize\textit{Open-source models for single image as input}}    \\ \noalign{\vskip 1pt}
MiniCPM-Llama3-V-2.5 & 86.49 & 53.33 & 75.61 & 73.15 & 64.00 & 43.48 & 31.91 & 46.85 & 50.00 & 50.00 & 50.00 \\ 
GLM-4V-9B & 78.38 & 53.33 & 75.61 & 70.37 & 60.00 & 45.65 & 31.91 & 46.15 & 61.76 & 55.00 & 58.11 \\
InternVL-Vicuna-13B & 72.97 & 43.33 & 85.37 & 69.44 & 68.00 & 58.70 & 29.79 & 52.45 & 73.53 & 72.50 & 72.97 \\
LLaVA-1.6 7B & 91.89 & \textbf{66.67} & 87.80 & \textbf{83.33} & 76.00 & 60.87 & 44.68 & 60.84 & 79.41 & 65.00 & 71.62 \\
InternVL-Vicuna-7B &86.49 & \textbf{66.67} & 82.93 & 79.63 & 64.00 & 65.22 & 42.55 & 57.34 & 67.65 & 52.50 & 59.46 \\
LLaVA-1.6 13B  & 94.59 & 56.67 & 90.24 & 82.41 & 78.00 & 69.57 & 36.17 & 61.54 & 64.71 & \textbf{77.50} & 71.62 \\
InternVL-Vicuna-13B-448px & 48.65 & 53.33 & 63.41 & 55.56 & 74.00 & 58.70 & 36.17 & 56.64 & 64.71 & 62.50 & 63.51 \\
mPLUG-Owl2-7B  & 91.89 & 60.00 & 90.24 & 82.41 & \textbf{84.00} & \textbf{71.74} & \textbf{51.06} & \textbf{69.23} & 70.59 & 70.00 & 70.27 \\
Mini-Gemini-7B-HD & 62.16 & 26.67 & 26.83 & 38.89 & 26.00 & 43.48 & 27.66 & 32.17 & 44.12 & 25.00 & 33.78 \\
SEAL & 70.27 & 46.67 & 63.41 & 61.11 & 64.00 & 50.00 & 44.68 & 53.15 & 41.18 & 55.00 & 48.65 \\
Mini-Gemini-7B & 37.84 & 16.67 & 21.95 & 25.93 & 6.00 & 17.39 & 8.51 & 10.49 & 20.59 & 20.00 & 20.27 \\
\bottomrule
\end{tabular}
\caption{Results on all sub-classes of zooming evaluation.} 
\label{tab:zoom}
\vspace{4em}
\end{table*}

\begin{table*}[h!]
\centering\scriptsize
\begin{tabular}{@{\hspace{0.05cm}}l@{\hspace{0.05cm}}|c@{\hspace{0.05cm}}c@{\hspace{0.05cm}}c@{\hspace{0.05cm}}c|c@{\hspace{0.05cm}}c@{\hspace{0.05cm}}c@{\hspace{0.1cm}}c|c@{\hspace{0.05cm}}c@{\hspace{0.1cm}}c@{\hspace{0.05cm}}}
\toprule
\multirow{2}{*}[-1ex]{Models} & \multicolumn{3}{c}{Type I} & \multirow{2}{*}[-1ex]{AVG} & \multicolumn{3}{c}{Type II} & \multirow{2}{*}[-1ex]{AVG} & \multicolumn{2}{c}{Type III} & \multirow{2}{*}[-1ex]{AVG} \\ \cmidrule(lr){2-4} \cmidrule(lr){6-8} \cmidrule(lr){10-11} 
& Geo-Loc & Orient & Daily-Loc &  & Obj-Attr & Obj-Rel & Count-Dis & & Event-M & Event-S & \\ \midrule\noalign{\vskip -3pt}
\multicolumn{12}{c}{\cellcolor[HTML]{EFEFEF} \scriptsize\textit{proprietary models}}    \\ \noalign{\vskip 1pt}
Gemini-1.5-pro & 91.89 & 50.00 & 82.93 & 76.85 & \textbf{76.00} & 50.00 & \textbf{55.32} & \textbf{60.84} & 70.59 & 57.50 & 63.51 \\
GPT-4o & \textbf{94.59} & \textbf{63.33} & 80.49 & \textbf{80.56} & 74.00 & 50.00 & 42.55 & 55.94 & \textbf{73.53} & \textbf{67.50} & \textbf{70.27} \\
Claude 3.5 Sonnet &91.89 & 53.33 & 80.49 & 76.85 & 72.00 & 52.17 & 40.43 & 55.24 & 64.71 & \textbf{67.50} & 66.22 \\
\midrule\noalign{\vskip -3pt}
\multicolumn{12}{c}{\cellcolor[HTML]{EFEFEF} \scriptsize\textit{Open-source models for multiple images as input}}    \\ \noalign{\vskip 1pt}
Qwen2-VL  & 91.89 & 50.00 & \textbf{85.37} & 77.78 & 72.00 & 54.35 & 38.30 & 55.24 & 52.94 & 45.00 & 48.65 \\
Idefics3-8B-Llama3  & 89.19 & 53.33 & \textbf{85.37} & 77.78 & 64.00 & 50.00 & 42.55 & 52.45 & 61.76 & 52.50 & 56.76 \\
MiniCPM-V 2.6  & 89.19 & 53.33 & 73.17 & 73.15 & 64.00 & 47.83 & 25.53 & 46.15 & 47.06 & 42.50 & 44.59 \\
mPLUG-Owl3  & 81.08 & 43.33 & 73.17 & 67.59 & 70.00 & 34.78 & 19.15 & 41.96 & 55.88 & 40.00 & 47.30 \\
LLaVA-OneVision  & 62.16 & 46.67 & 73.17 & 62.04 & 64.00 & 52.17 & 23.40 & 46.85 & 61.76 & 47.50 & 54.05 \\
InternVL2-8B &78.38 & 50.00 & 80.49 & 71.30 & 62.00 & 41.30 & 31.91 & 45.45 & 35.29 & 60.00 & 48.65 \\
Mantis  & 91.89 & 40.00 & 70.73 & 69.44 & 70.00 & 50.00 & 19.15 & 46.85 & 52.94 & 40.00 & 45.95 \\
Idefics2-8B &75.68 & 60.00 & 70.73 & 69.44 & 60.00 & 39.13 & 19.15 & 39.86 & 61.76 & 52.50 & 56.76 \\	
Brote-IM-XL-3B  & 70.27 & 43.33 & 65.85 & 61.11 & 62.00 & 41.30 & 42.55 & 48.95 & 47.06 & 35.00 & 40.54 \\
Idefics2-8B-base &86.49 & 43.33 & 78.05 & 71.30 & 54.00 & 36.96 & 23.40 & 38.46 & 50.00 & 27.50 & 37.84 \\
Brote-IM-XXL-11B  & 70.27 & 40.00 & 65.85 & 60.19 & 56.00 & 47.83 & 31.91 & 45.45 & 55.88 & 32.50 & 43.24 \\
MMICL-XXL-11B  & 62.16 & 53.33 & 63.41 & 60.19 & 56.00 & 47.83 & 34.04 & 46.15 & 52.94 & 32.50 & 41.89 \\
MMICL-XL-3B   & 32.43 & 50.00 & 65.85 & 50.00 & 52.00 & 45.65 & 38.30 & 45.45 & 41.18 & 32.50 & 36.49 \\
\midrule\noalign{\vskip -3pt}
\multicolumn{12}{c}{\cellcolor[HTML]{EFEFEF} \scriptsize\textit{Open-source models for single image as input}}    \\ \noalign{\vskip 1pt}
MiniCPM-Llama3-V-2.5  & \textbf{94.59} & 36.67 & 82.93 & 74.07 & 66.00 & 50.00 & 48.94 & 55.24 & 41.18 & 57.50 & 50.00 \\ 
GLM-4V-9B &86.49 & 53.33 & 80.49 & 75.00 & 62.00 & 30.43 & 40.43 & 44.76 & 55.88 & 52.50 & 54.05 \\
InternVL-Vicuna-13B &62.16 & 46.67 & 60.98 & 57.41 & 64.00 & 54.35 & 25.53 & 48.25 & 58.82 & 60.00 & 59.46 \\
InternVL-Vicuna-7B &72.97 & 50.00 & 60.98 & 62.04 & 60.00 & 45.65 & 34.04 & 46.85 & 55.88 & 47.50 & 51.35 \\
InternVL-Vicuna-13B-448px &45.95 & 40.00 & 56.10 & 48.15 & 62.00 & \textbf{56.52} & 25.53 & 48.25 & 50.00 & 47.50 & 48.65 \\
mPLUG-Owl2-7B  & 64.86 & 40.00 & 53.66 & 53.70 & 60.00 & 47.83 & 19.15 & 42.66 & 52.94 & 50.00 & 51.35 \\
Mini-Gemini-7B-HD &72.97 & 53.33 & 43.90 & 56.48 & 56.00 & 43.48 & 25.53 & 41.96 & 58.82 & 42.50 & 50.00 \\
SEAL & 56.76 & 43.33 & 53.66 & 51.85 & 54.00 & 41.30 & 19.15 & 38.46 & 29.41 & 40.00 & 35.14 \\
Mini-Gemini-7B &59.46 & 46.67 & 43.90 & 50.00 & 36.00 & 39.13 & 29.79 & 34.97 & 32.35 & 32.50 & 32.43 \\
\bottomrule
\end{tabular}
\caption{Results on each sub-classes of Shift-R, shifting with random initial views.} 
\label{tab:shiftR}
\end{table*}

\begin{table*}[h!]
\centering\scriptsize
\begin{tabular}{@{\hspace{0.05cm}}l@{\hspace{0.05cm}}|c@{\hspace{0.05cm}}c@{\hspace{0.05cm}}c@{\hspace{0.05cm}}c|c@{\hspace{0.05cm}}c@{\hspace{0.05cm}}c@{\hspace{0.1cm}}c|c@{\hspace{0.05cm}}c@{\hspace{0.1cm}}c@{\hspace{0.05cm}}}
\toprule
\multirow{2}{*}[-1ex]{Models} & \multicolumn{3}{c}{Type I} & \multirow{2}{*}[-1ex]{AVG} & \multicolumn{3}{c}{Type II} & \multirow{2}{*}[-1ex]{AVG} & \multicolumn{2}{c}{Type III} & \multirow{2}{*}[-1ex]{AVG} \\ \cmidrule(lr){2-4} \cmidrule(lr){6-8} \cmidrule(lr){10-11} 
& Geo-Loc & Orient & Daily-Loc &  & Obj-Attr & Obj-Rel & Count-Dis & & Event-M & Event-S & \\ \midrule\noalign{\vskip -3pt}
\multicolumn{12}{c}{\cellcolor[HTML]{EFEFEF} \scriptsize\textit{proprietary models}}    \\ \noalign{\vskip 1pt}
Gemini-1.5-pro  & 91.89 & 63.33 & \textbf{90.24} & \textbf{83.33} & 68.00 & 50.00 & \textbf{48.94} & \textbf{55.94} & \textbf{79.41} & 52.50 & 64.86 \\
GPT-4o & \textbf{97.30} & 53.33 & 85.37 & 80.56 & 64.00 & 52.17 & 44.68 & 53.85 & 73.53 & \textbf{72.50} & \textbf{72.97} \\
Claude 3.5 Sonnet &94.59 & \textbf{70.00} & 80.49 & 82.41 & 66.00 & 52.17 & 38.30 & 52.45 & 70.59 & 65.00 & 67.57 \\
\midrule\noalign{\vskip -3pt}
\multicolumn{12}{c}{\cellcolor[HTML]{EFEFEF} \scriptsize\textit{Open-source models for multiple images as input}}    \\ \noalign{\vskip 1pt}
Qwen2-VL  & 94.59 & 53.33 & 85.37 & 79.63 & 68.00 & 60.87 & 40.43 & 56.64 & 50.00 & 50.00 & 50.00 \\
Idefics3-8B-Llama3  & 89.19 & 46.67 & 82.93 & 75.00 & 60.00 & 56.52 & 40.43 & 52.45 & 58.82 & 42.50 & 50.00\\
MiniCPM-V 2.6  & 89.19 & 63.33 & 78.05 & 77.78 & \textbf{76.00} & 58.70 & 23.40 & 53.15 & 52.94 & 52.50 & 52.70 \\
mPLUG-Owl3  & 83.78 & 46.67 & 78.05 & 71.30 & 62.00 & 52.17 & 27.66 & 47.55 & 52.94 & 50.00 & 51.35\\
LLaVA-OneVision  & 70.27 & 43.33 & 70.73 & 62.96 & \textbf{76.00} & 60.87 & 29.79 & 55.94 & 58.82 & 45.00 & 51.35 \\
InternVL2-8B &83.78 & 63.33 & 63.41 & 70.37 & 66.00 & 56.52 & 36.17 & 53.15 & 44.12 & 67.50 & 56.76 \\
Mantis  & 91.89 & 36.67 & 70.73 & 68.52 & 72.00 & 52.17 & 23.40 & 49.65 & 50.00 & 45.00 & 47.30 \\
Idefics2-8B &83.78 & 56.67 & 78.05 & 74.07 & 68.00 & 43.48 & 25.53 & 46.15 & 52.94 & 55.00 & 54.05 \\	
Brote-IM-XL-3B  & 62.16 & 40.00 & 68.29 & 58.33 & 64.00 & 50.00 & 44.68 & 53.15 & 41.18 & 45.00 & 43.24\\
Idefics2-8B-base &81.08 & 43.33 & 85.37 & 72.22 & 62.00 & 39.13 & 25.53 & 42.66 & 41.18 & 27.50 & 33.78 \\
Brote-IM-XXL-11B  & 64.86 & 33.33 & 60.98 & 54.63 & 58.00 & 52.17 & 46.81 & 52.45 & 41.18 & 42.50 & 41.89 \\
MMICL-XXL-11B  & 62.16 & 36.67 & 60.98 & 54.63 & 62.00 & 41.30 & 46.81 & 50.35 & 41.18 & 42.50 & 41.89
 \\
MMICL-XL-3B  & 56.76 & 46.67 & 68.29 & 58.33 & 56.00 & 45.65 & 40.43 & 47.55 & 41.18 & 35.00 & 37.84 \\
\midrule\noalign{\vskip -3pt}
\multicolumn{12}{c}{\cellcolor[HTML]{EFEFEF} \scriptsize\textit{Open-source models for single image as input}}    \\ \noalign{\vskip 1pt}
MiniCPM-Llama3-V-2.5  & 91.89 & 46.67 & 80.49 & 75.00 & 58.00 & 45.65 & 44.68 & 49.65 & 47.06 & 57.50 & 52.70 \\ 
GLM-4V-9B &94.59 & 56.67 & 78.05 & 77.78 & 66.00 & 47.83 & 36.17 & 50.35 & 52.94 & 57.50 & 55.41 \\
InternVL-Vicuna-13B &59.46 & 43.33 & 65.85 & 57.41 & 60.00 & \textbf{63.04} & 25.53 & 49.65 & 52.94 & 52.50 & 52.70 \\
InternVL-Vicuna-7B & 64.86 & 53.33 & 58.54 & 59.26 & 62.00 & 45.65 & 29.79 & 46.15 & 55.88 & 45.00 & 50.00 \\
LLaVA-1.6 13B  & 70.27 & 46.67 & 68.29 & 62.96 & 72.00 & 43.48 & 29.79 & 48.95 & 50.00 & 67.50 & 59.46 \\
InternVL-Vicuna-13B-448px &56.76 & 40.00 & 51.22 & 50.00 & 54.00 & 58.70 & 27.66 & 46.85 & 50.00 & 47.50 & 48.65 \\
mPLUG-Owl2-7B  & 67.57 & 43.33 & 56.10 & 56.48 & 50.00 & 47.83 & 23.40 & 40.56 & 47.06 & 47.50 & 47.30 \\
Mini-Gemini-7B-HD &67.57 & 53.33 & 51.22 & 57.41 & 52.00 & 43.48 & 29.79 & 41.96 & 52.94 & 40.00 & 45.95 \\
SEAL &56.76 & 43.33 & 53.66 & 51.85 & 52.00 & 36.96 & 25.53 & 38.46 & 29.41 & 45.00 & 37.84 \\
Mini-Gemini-7B &62.16 & 36.67 & 36.59 & 45.37 & 40.00 & 45.65 & 19.15 & 34.97 & 32.35 & 35.00 & 33.78 \\
\bottomrule
\end{tabular}
\caption{Results on sub-classes of Shift-E (the easy-level shifting evaluation), where initial views contain clues for answering the question.} 
\label{tab:shift_easy}
\end{table*}

\begin{table*}[h!]
\centering\scriptsize
\begin{tabular}{@{\hspace{0.05cm}}l@{\hspace{0.05cm}}|c@{\hspace{0.05cm}}c@{\hspace{0.05cm}}c@{\hspace{0.05cm}}c|c@{\hspace{0.05cm}}c@{\hspace{0.05cm}}c@{\hspace{0.1cm}}c|c@{\hspace{0.05cm}}c@{\hspace{0.1cm}}c@{\hspace{0.05cm}}}
\toprule
\multirow{2}{*}[-1ex]{Models} & \multicolumn{3}{c}{Type I} & \multirow{2}{*}[-1ex]{AVG} & \multicolumn{3}{c}{Type II} & \multirow{2}{*}[-1ex]{AVG} & \multicolumn{2}{c}{Type III} & \multirow{2}{*}[-1ex]{AVG} \\ \cmidrule(lr){2-4} \cmidrule(lr){6-8} \cmidrule(lr){10-11} 
& Geo-Loc & Orient & Daily-Loc &  & Obj-Attr & Obj-Rel & Count-Dis & & Event-M & Event-S & \\ \midrule\noalign{\vskip -3pt}
\multicolumn{12}{c}{\cellcolor[HTML]{EFEFEF} \scriptsize\textit{proprietary models}}    \\ \noalign{\vskip 1pt}
Gemini-1.5-pro  & 89.19 & 60.00 & \textbf{90.24} & \textbf{81.48} & 70.00 & 47.83 & 42.55 & 53.85 & \textbf{73.53} & 60.00 & 66.22 \\
GPT-4o & \textbf{94.59} & 53.33 & 85.37 & 79.63 & 64.00 & 52.17 & 42.55 & 53.15 & 67.65 & \textbf{70.00} & \textbf{68.92} \\
Claude 3.5 Sonnet &89.19 & 46.67 & 75.61 & 72.22 & 66.00 & 54.35 & 40.43 & 53.85 & 55.88 & 52.50 & 54.05 \\
\midrule\noalign{\vskip -3pt}
\multicolumn{12}{c}{\cellcolor[HTML]{EFEFEF} \scriptsize\textit{Open-source models for multiple images as input}}    \\ \noalign{\vskip 1pt}
Qwen2-VL  & 91.89 & 50.00 & 85.37 & 77.78 & \textbf{76.00} & \textbf{60.87} & 40.43 & \textbf{59.44} & 55.88 & 52.50 & 54.05 \\
Idefics3-8B-Llama3  & 86.49 & 50.00 & 80.49 & 74.07 & 64.00 & 52.17 & 40.43 & 52.45 & 55.88 & 50.00 & 52.70\\
MiniCPM-V 2.6  & 86.49 & \textbf{63.33} & 73.17 & 75.00 & 62.00 & 56.52 & 21.28 & 46.85 & 55.88 & 55.00 & 55.41 \\
mPLUG-Owl3  & 81.08 & 46.67 & 68.29 & 66.67 & 62.00 & 54.35 & 25.53 & 47.55 & 58.82 & 52.50 & 55.41\\
LLaVA-OneVision  & 56.76 & 46.67 & 63.41 & 56.48 & 62.00 & 56.52 & 27.66 & 48.95 & 64.71 & 42.50 & 52.70 \\
InternVL2-8B &78.38 & 50.00 & 63.41 & 64.81 & 66.00 & 43.48 & 31.91 & 47.55 & 44.12 & 50.00 & 47.30 \\
Mantis  & 89.19 & 36.67 & 65.85 & 65.74 & 62.00 & 54.35 & 19.15 & 45.45 & 55.88 & 42.50 & 48.65 \\
Idefics2-8B & 67.57 & \textbf{63.33} & 70.73 & 67.59 & 56.00 & 43.48 & 21.28 & 40.56 & 52.94 & 50.00 & 51.35\\	
Brote-IM-XL-3B  & 48.65 & 36.67 & 63.41 & 50.93 & 54.00 & 50.00 & 44.68 & 49.65 & 44.12 & 35.00 & 39.19 \\
Idefics2-8B-base &75.68 & 43.33 & 82.93 & 69.44 & 52.00 & 41.30 & 21.28 & 38.46 & 41.18 & 25.00 & 32.43 \\
Brote-IM-XXL-11B  & 56.76 & 30.00 & 63.41 & 51.85 & 42.00 & 41.30 & 42.55 & 41.96 & 44.12 & 37.50 & 40.54\\
MMICL-XXL-11B  & 56.76 & 40.00 & 63.41 & 54.63 & 50.00 & 41.30 & 42.55 & 44.76 & 44.12 & 35.00 & 39.19 \\
MMICL-XL-3B   & 40.54 & 43.33 & 68.29 & 51.85 & 48.00 & 47.83 & 40.43 & 45.45 & 44.12 & 32.50 & 37.84 \\
\midrule\noalign{\vskip -3pt}
\multicolumn{12}{c}{\cellcolor[HTML]{EFEFEF} \scriptsize\textit{Open-source models for single image as input}}    \\ \noalign{\vskip 1pt}
MiniCPM-Llama3-V-2.5  & 91.89 & 46.67 & 73.17 & 72.22 & 58.00 & 43.48 & \textbf{46.81} & 49.65 & 47.06 & 57.50 & 52.70 \\ 
GLM-4V-9B &89.19 & 56.67 & 73.17 & 74.07 & 50.00 & 43.48 & 38.30 & 44.06 & 55.88 & 50.00 & 52.70 \\
InternVL-Vicuna-13B &67.57 & 40.00 & 65.85 & 59.26 & 56.00 & 56.52 & 23.40 & 45.45 & 55.88 & 60.00 & 58.11 \\
InternVL-Vicuna-7B &62.16 & 43.33 & 63.41 & 57.41 & 62.00 & 45.65 & 25.53 & 44.76 & 52.94 & 52.50 & 52.70 \\
LLaVA-1.6 13B  & 62.16 & 53.33 & 65.85 & 61.11 & 62.00 & 50.00 & 29.79 & 47.55 & 52.94 & 50.00 & 51.35 \\
InternVL-Vicuna-13B-448px &43.24 & 46.67 & 53.66 & 48.15 & 60.00 & 50.00 & 27.66 & 46.15 & 50.00 & 57.50 & 54.05 \\
mPLUG-Owl2-7B  & 59.46 & 40.00 & 58.54 & 53.70 & 54.00 & 50.00 & 23.40 & 42.66 & 47.06 & 47.50 & 47.30 \\
Mini-Gemini-7B-HD &62.16 & 56.67 & 39.02 & 51.85 & 56.00 & 47.83 & 25.53 & 43.36 & 58.82 & 42.50 & 50.00 \\
SEAL &56.76 & 43.33 & 48.78 & 50.00 & 52.00 & 34.78 & 23.40 & 37.06 & 26.47 & 40.00 & 33.78 \\
Mini-Gemini-7B &72.97 & 36.67 & 43.90 & 51.85 & 34.00 & 39.13 & 21.28 & 31.47 & 35.29 & 27.50 & 31.08 \\
\bottomrule
\end{tabular}
\caption{Results on each sub-classes of Shift-M (the medium-level shifting evaluation), where initial views contain only partial clues for answering the questions.}
\label{tab:shift_medium}
\end{table*}

\begin{table*}[h!]
\centering\scriptsize
\begin{tabular}{@{\hspace{0.05cm}}l@{\hspace{0.05cm}}|c@{\hspace{0.05cm}}c@{\hspace{0.05cm}}c@{\hspace{0.05cm}}c|c@{\hspace{0.05cm}}c@{\hspace{0.05cm}}c@{\hspace{0.1cm}}c|c@{\hspace{0.05cm}}c@{\hspace{0.1cm}}c@{\hspace{0.05cm}}}
\toprule
\multirow{2}{*}[-1ex]{Models} & \multicolumn{3}{c}{Type I} & \multirow{2}{*}[-1ex]{AVG} & \multicolumn{3}{c}{Type II} & \multirow{2}{*}[-1ex]{AVG} & \multicolumn{2}{c}{Type III} & \multirow{2}{*}[-1ex]{AVG} \\ \cmidrule(lr){2-4} \cmidrule(lr){6-8} \cmidrule(lr){10-11} 
& Geo-Loc & Orient & Daily-Loc &  & Obj-Attr & Obj-Rel & Count-Dis & & Event-M & Event-S & \\ \midrule\noalign{\vskip -3pt}
\multicolumn{12}{c}{\cellcolor[HTML]{EFEFEF} \scriptsize\textit{proprietary models}}    \\ \noalign{\vskip 1pt}
Gemini-1.5-pro  & \textbf{91.89} & \textbf{66.67} & \textbf{90.24} & \textbf{84.26} & 70.00 & 50.00 & \textbf{46.81} & 55.94 & 73.53 & 57.50 & 64.86 \\
GPT-4o & \textbf{91.89} & 53.33 & 82.93 & 77.78 & 66.00 & 47.83 & 34.04 & 49.65 & \textbf{79.41} & \textbf{65.00} & \textbf{71.62}\\
Claude 3.5 Sonnet & \textbf{91.89} & 46.67 & 73.17 & 72.22 & 68.00 & 52.17 & 34.04 & 51.75 & 67.65 & 62.50 & 64.86 \\
\midrule\noalign{\vskip -3pt}
\multicolumn{12}{c}{\cellcolor[HTML]{EFEFEF} \scriptsize\textit{Open-source models for multiple images as input}}    \\ \noalign{\vskip 1pt}
Qwen2-VL  & \textbf{91.89} & 50.00 & 82.93 & 76.85 & \textbf{78.00} & 52.17 & 40.43 & \textbf{57.34} & 50.00 & 47.50 & 48.65\\
Idefics3-8B-Llama3  & 83.78 & 53.33 & 85.37 & 75.93 & 68.00 & 52.17 & 36.17 & 52.45 & 58.82 & 47.50 & 52.70 \\
MiniCPM-V 2.6  & 86.49 & 50.00 & 75.61 & 72.22 & 64.00 & 50.00 & 23.40 & 46.15 & 52.94 & 47.50 & 50.00 \\
mPLUG-Owl3  & 78.38 & 43.33 & 70.73 & 65.74 & 54.00 & 50.00 & 25.53 & 43.36 & 58.82 & 52.50 & 55.41 \\
InternVL2-8B &64.86 & 40.00 & 65.85 & 58.33 & 62.00 & \textbf{54.35} & 34.04 & 50.35 & 52.94 & 40.00 & 45.95\\
LLaVA-OneVision  & 56.76 & 40.00 & 63.41 & 54.63 & 54.00 & 52.17 & 27.66 & 44.76 & 55.88 & 40.00 & 47.30 \\
Mantis  & 86.49 & 36.67 & 63.41 & 63.89 & 66.00 & 50.00 & 19.15 & 45.45 & 58.82 & 40.00 & 48.65 \\
Idefics2-8B &62.16 & \textbf{66.67} & 65.85 & 64.81 & 52.00 & 41.30 & 23.40 & 39.16 & 52.94 & 42.50 & 47.30\\	
Brote-IM-XL-3B  & 54.05 & 43.33 & 58.54 & 52.78 & 52.00 & 36.96 & 40.43 & 43.36 & 50.00 & 35.00 & 41.89 \\
Idefics2-8B-base &81.08 & 46.67 & 75.61 & 69.44 & 50.00 & 43.48 & 19.15 & 37.76 & 41.18 & 27.50 & 33.78\\
Brote-IM-XXL-11B  & 56.76 & 30.00 & 60.98 & 50.93 & 44.00 & 34.78 & 38.30 & 39.16 & 50.00 & 35.00 & 41.89 \\
MMICL-XXL-11B  & 64.86 & 46.67 & 58.54 & 57.41 & 50.00 & 30.43 & 36.17 & 39.16 & 50.00 & 32.50 & 40.54 \\
MMICL-XL-3B   & 43.24 & 43.33 & 68.29 & 52.78 & 46.00 & 32.61 & 38.30 & 39.16 & 52.94 & 32.50 & 41.89 \\
\midrule\noalign{\vskip -3pt}
\multicolumn{12}{c}{\cellcolor[HTML]{EFEFEF} \scriptsize\textit{Open-source models for single image as input}}    \\ \noalign{\vskip 1pt}
MiniCPM-Llama3-V-2.5  & \textbf{91.89} & 36.67 & 78.05 & 71.30 & 60.00 & 45.65 & 42.55 & 49.65 & 50.00 & 55.00 & 52.70  \\ 
GLM-4V-9B &89.19 & 50.00 & 73.17 & 72.22 & 40.00 & 34.78 & 38.30 & 37.76 & 58.82 & 50.00 & 54.05\\
InternVL-Vicuna-13B &56.76 & 36.67 & 60.98 & 52.78 & 62.00 & 50.00 & 21.28 & 44.76 & 61.76 & 60.00 & 60.81\\
InternVL-Vicuna-7B &59.46 & 43.33 & 63.41 & 56.48 & 52.00 & 45.65 & 23.40 & 40.56 & 55.88 & 50.00 & 52.70\\
LLaVA-1.6 13B  & 51.35 & 50.00 & 60.98 & 54.63 & 58.00 & 41.30 & 29.79 & 43.36 & 55.88 & 52.50 & 54.05 \\
InternVL-Vicuna-13B-448px &48.65 & 40.00 & 60.98 & 50.93 & 58.00 & 47.83 & 29.79 & 45.45 & 50.00 & 52.50 & 51.35\\
mPLUG-Owl2-7B  & 56.76 & 40.00 & 56.10 & 51.85 & 58.00 & 45.65 & 25.53 & 43.36 & 52.94 & 42.50 & 47.30 \\
Mini-Gemini-7B-HD &70.27 & 50.00 & 51.22 & 57.41 & 52.00 & 47.83 & 29.79 & 43.36 & 58.82 & 47.50 & 52.70\\
SEAL  &56.76 & 36.67 & 51.22 & 49.07 & 54.00 & 34.78 & 23.40 & 37.76 & 29.41 & 37.50 & 33.78 \\
Mini-Gemini-7B &64.86 & 43.33 & 51.22 & 53.70 & 36.00 & 28.26 & 23.40 & 29.37 & 26.47 & 20.00 & 22.97 \\
\bottomrule
\end{tabular}
\vspace{-1em}
\caption{Results on each sub-classes of Shift-H (the hard-level shifting evaluation), where initial views do not display clues for answering the questions. Models should decide whether to shift to the next view all by themselves.} 
\label{tab:shift_hard}
\end{table*}

\subsection{Analysis of Results on Zooming Evaluation}
We notice that for the zooming evaluation, except for InternVL and LLaVA-1.6, single-image models fail to achieve equivalent or comparable results (comparing with full image setting), and present performance gap of as large as 29.23\% (for Mini-Gemini-7B) where the zooming results are much lower. These imply that some single models are unaware of the location of key visual information required by the target question. On the contrary, multi-image models present comparable or even better scores under the zooming evaluation.

We summarise the zooming results on sub-classes from Table~\ref{tab:zoom}, that the environment-centric category (including Geo-Loc, Orient, and Daily-Loc) presents significantly higher scores than object-centric and event-centric categories. The reason lies in the fact that questions in environment-centric category require more visual commonsense that most of models learnt from the vast training data. We also notice that Idefics2-8B-base even enlarges the performance gap between environment-centric category and the others by around 40\%, which demonstrate extremely unbalanced capabilities of exploiting inherent commonsense and observed visual clues.  
The most challenging types of instances are Orient, Count-Dis and Event-S, that present even halved scores compared to the other sub-classes.
Surprisingly, some of evaluated single-image models achieve better scores or perform equally compared to powerful proprietary models for the zooming evaluation, especially mPLUG-Owl2-7B regarding the object-centric category. We hypothesis that this model possesses strong object recognition ability and is less affected by object hallucination compared to other MLLMs.

\subsection{Analysis of Results on Shifting Evaluation} \label{app:shift}
The shifting pipeline aims at mimicking the scenario when humans look for more visual information by shifting the perceptual fields, the previously perceived views cannot be simply erased from memory, and new views are integrated incrementally. 
Results of Shift-R evaluation are shown in Table~\ref{tab:shiftR}, and the level-specified shifting evaluation are listed in Table~\ref{tab:shift_easy}, Table~\ref{tab:shift_medium} and Table~\ref{tab:shift_hard}. Similar to that of zooming evaluation, results on environment-centric category are significantly better than the ones on object-centric and event-centric categories. The results of proprietary models are better than the results of open-source models, and that models for multiple images perform better than models for single image. 
We observe a trend where, as the difficulty increases, the superiority of open-source multi-image models becomes more evident.

There is an overall trend for all the sub-classed that the accuracy decreases as the difficulty is getting increased. As shown by typical results for LLaVA-OneVision and GLM-4V-9B in Table~\ref{tab:rst_}, the gaps between Shift-E and Shift-H are as large as 8.61\% and 7.70\%, respectively. However, exceptional performances are identified for Gemini-1.5-pro, Idefics3-8B-Llama3, and Mini-Gemini-7B-HD, where the results of Shift-H even outperform the results of Shift-E. One of the reason could be the recall of selected views. For Gemini-1.5-pro, the recall for Shift-H is 47.73\%, over 1 point higher than Shift-E (45.29\%). We conclude that Gemini-1.5-pro achieves higher accuracy on Shift-H due to the acquisition of more proper views. While Idefics3 presents a different trend. It maintains a recall of 74.64\% from Shift-E to Shift-H, but achieves a higher accuracy on Shift-H. There is another potential reason that the performance gain of this model comes from the order of input views. The hard-level evaluation starts with less relevant views and appends more useful views at the end of the image sequence, and the performance of these models are more significantly influenced by the order of presented images compared to the rest models. The degradation of performance is more remarkable for the environment-centric and the object-centric categories compared to the event-centric category. Regarding the increasing of difficulty for the environment-centric and the object-centric categories, we observe gaps of about 10\% for models such as LLaVA-OneVision, Idefics2-8B, Brote, MMICL, GLM-4V-9B and Mini-Gemini-7B. These observations indicate that different initial perceptual fields have distinct impacts on instances that requiring demanding attention on subtle changes of fine-grained objects.
Results show that GPT- 4o consistently outperforms other models in the average score of the environment-centric category, implying robust event capture and understanding capabilities in multi-image scenarios.

\subsection{Analysis of View Selection}\label{app:view_select}
Our evaluation pipelines involve selecting useful view in their first stages. The reliability of selected views plays a crucial role in the following question answering stage. We compute the recall of used views following Equation~\ref{eq:recall} in Appendix~\ref{app:view_measure}, and include results in Table~\ref{tab:gt_rst}, along with accuracy of providing models with groundtruth views. Overall, lower selection recall tends to correlate with lower question answering accuracy. For example, Idefics3-8B-Llama3 and InternVL2-8B present the lowest recalls (41.09\%) among multi-image models in Table~\ref{tab:gt_rst}, leading to lower zooming evaluation accuracies of 56.00\% and 58.15\%, respectively. 

For zooming evaluation, we also investigate the performance when the given groundtruth views that contain human-annotated clues. Generally, models are prompted to generate more accurate answers compared to the pure zooming setting. However, mPLUG-Owl3, Gemini-1.5-pro, and LLaVA-OneVision are only exceptional, whose performance slightly degrade when given the visual clues. We argue that these models are better at the question answering task rather than exhibiting active perception ability. Additionally, we observe that shifting evaluations tend to require more views to be used for answering questions than zooming evaluations, yet it often results in inferior overall performance compared to zooming. For the shifting evaluation, some models keep shifting view until all four views are inquired. However, this does not necessarily support a better accuracy, as some views contain redundant information that might distract the model during reasoning. 
Additionally, we observe shifting evaluations tend to require more views to be used for answering questions than zooming evaluations, yet it often results in inferior overall performance compared to zooming.
This is because some of the current advanced models struggle to either move their field of views for necessary visual details, or screen out distracting information. Therefore, we believe that more attention should be paid to evaluating and enhancing active perception abilities of MLLMs given constraint perceptual fields.

\begin{table}[t]
    \centering\scriptsize
        \begin{tabular}{@{\hspace{0.05cm}}l@{\hspace{0.05cm}}|@{\hspace{0.05cm}}c@{\hspace{0.05cm}}|@{\hspace{0.05cm}}c@{\hspace{0.05cm}}c@{\hspace{0.05cm}}c@{\hspace{0.05cm}}|@{\hspace{0.05cm}}c@{\hspace{0.05cm}}c@{\hspace{0.05cm}}c@{\hspace{0.02cm}}}
        \toprule
        \multirow{2}{*}[-1ex]{Models} & \multirow{2}{*}[-1ex]{ACC$_{\text{GT}}$} & \multicolumn{3}{@{\hspace{0.05cm}}c@{\hspace{0.1cm}}|@{\hspace{0.1cm}}}{Zooming} & \multicolumn{3}{c}{Shift-R} \\ \cmidrule(lr){3-5} \cmidrule(lr){6-8} 
         & & ACC$_{\text{QA}}$ & $R_{\text{select}}$ & \#view & ACC$_{\text{QA}}$ & $R_{\text{select}}$ & \#view \\
        \midrule\noalign{\vskip -3pt}
        \multicolumn{8}{c}{\cellcolor[HTML]{EFEFEF} \scriptsize\textit{Multi-image Models}}    \\ \noalign{\vskip 1pt}
        GPT-4o & \underline{73.54} & 68.62 & \underline{69.03} & 2.29 & \textbf{67.08} & 60.54 & 3.26 \\
        mPLUG-Owl3 & 60.62 & 60.92 & 68.57 & 2.66 & 51.69 & \textbf{74.62} & 4.00 \\
        Claude 3.5 Sonnet & 72.31 & \underline{71.69} & 67.64 & 2.19 & \underline{65.23} & 45.52 & 2.47 \\
        Qwen2-VL & 65.85 & 64.62 & 64.61 & 2.35 & 61.23 & \textbf{74.62} & 4.00  \\
        Gemini-1.5-pro & 72.00 & \textbf{72.31} & 62.63 & 2.10 & \textbf{67.08} & 46.33 & 2.46 \\ 
        MiniCPM-V 2.6 & 62.77 & 61.85 & 57.03 & 2.20 & 54.77 & 54.83 & 2.98 \\
        LLaVA-OneVision & 64.92 & 65.23 & 46.67 & 2.35 & 53.54 & 37.14 & 2.02 \\
        Idefics3-8B-Llama3& 60.92 & 58.15 & 41.09 & 1.52 & 61.85 & \textbf{74.62} & 4.00 \\
        InternVL2-8B & 73.23 & 56.00 & 41.09 & 1.53 & 54.77 & 45.75 & 2.61  \\
        \midrule\noalign{\vskip -3pt}
        \multicolumn{8}{c}{\cellcolor[HTML]{EFEFEF} \scriptsize\textit{Single-image Models}}    \\ \noalign{\vskip 1pt}
        InternVL-Vicuna-13B & 68.00 & 62.77 & \textbf{83.47} & 3.31 & 53.85 & 69.73 & 3.75 \\
        LLaVA-1.6 13B & 67.69 & 68.92 & 68.57 & 2.65 & 51.69 & \textbf{74.62} & 4.00 \\
        SEAL  & 56.92 & 54.77 & 68.22 & 2.74 & 42.15 & \underline{71.48} & 3.83 \\
        MiniCPM-Llama3-V-2.5 & 62.20 & 61.25 & 66.12 & 2.46 & 53.85 & 63.56 & 3.42  \\
        mPLUG-Owl2-7B & 67.38 & 55.38 & 47.61 & 1.97 & 47.38 & \textbf{74.62} & 4.00 \\
        GLM-4V-9B & \textbf{74.46} & 56.92 & 30.62 & 1.08 & 56.92 & 50.17 & 2.64  \\
        \bottomrule
        \end{tabular}
        \caption{Results of view selection (``$R_{\text{select}}$'' ) and the accuracy given groundtruth views (``ACC$_{\text{GT}}$'', 2.64 views on average) that contain human-annotated visual clues. ``ACC$_{\text{QA}}$'': accuracy of question answering for zooming and shifting. ``\#view'': average counts of selected views.} 
        \label{tab:gt_rst}
\end{table}

\begin{figure*}
    \centering
    \includegraphics[width=1.\textwidth]{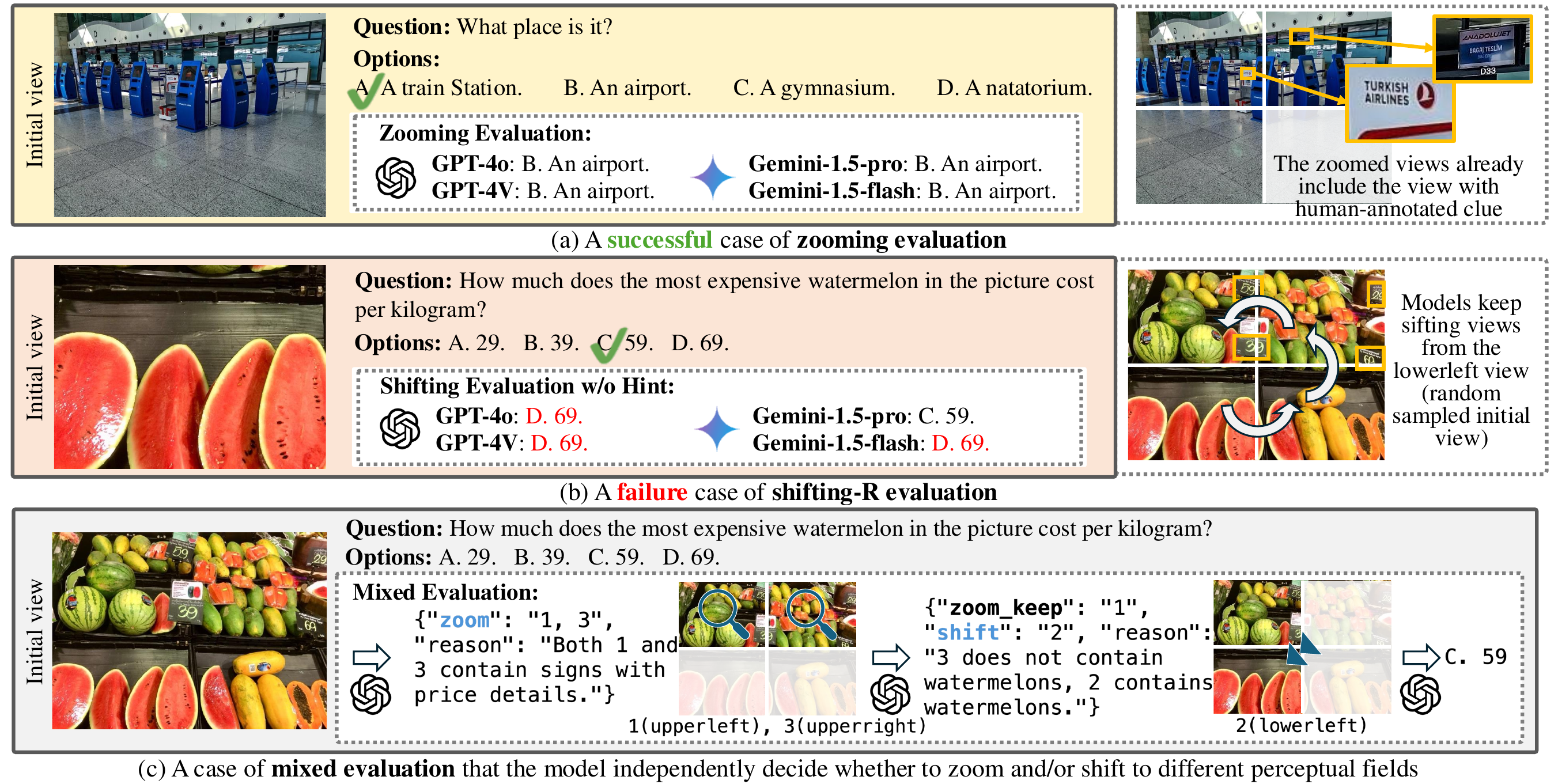}
    \vspace{-1em}
    \caption{Cases for each evaluation pipelines. (a) a succeeded \textbf{zooming} case, (b) a failed \textbf{shifting} case, and (c) a \textbf{mixed} case that successfully corrects the wrong answer produced by (b). Model selected views for case (a) and (b) are placed to the right of example frames, and used views for case (c) are shown with in its frame as the selection of views changes during the evaluation process.\label{fig:3cases}}
    \vspace{-1em}
\end{figure*}

\section{Case Study}\label{app:case_study}
\begin{figure*}
    \centering
    \begin{subfigure}{0.40\textwidth}
        \centering
        \includegraphics[height=4.0cm]{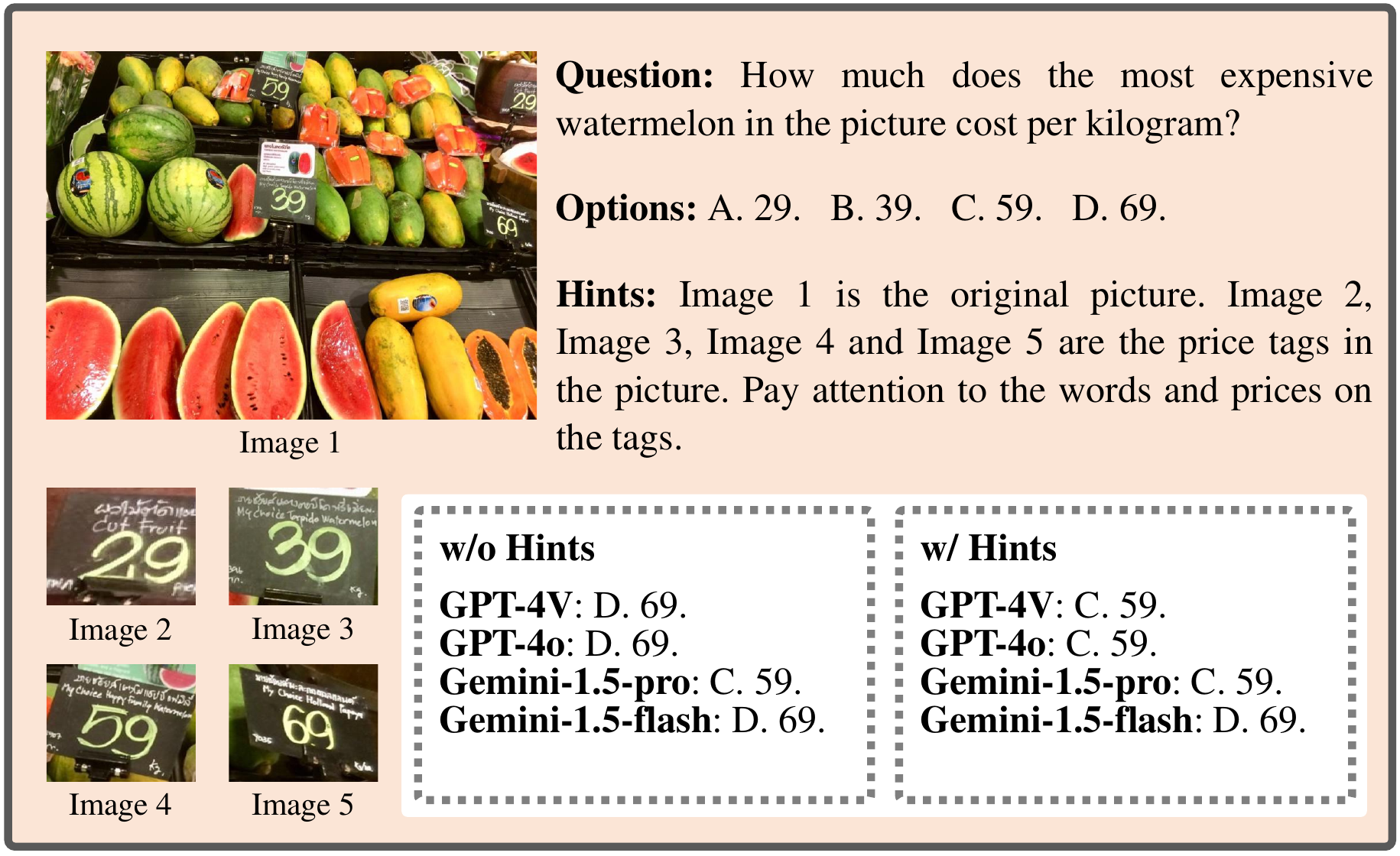}
    \end{subfigure}
    \hfill
    \begin{subfigure}{0.36\textwidth}
        \centering
        \includegraphics[height=4.0cm]{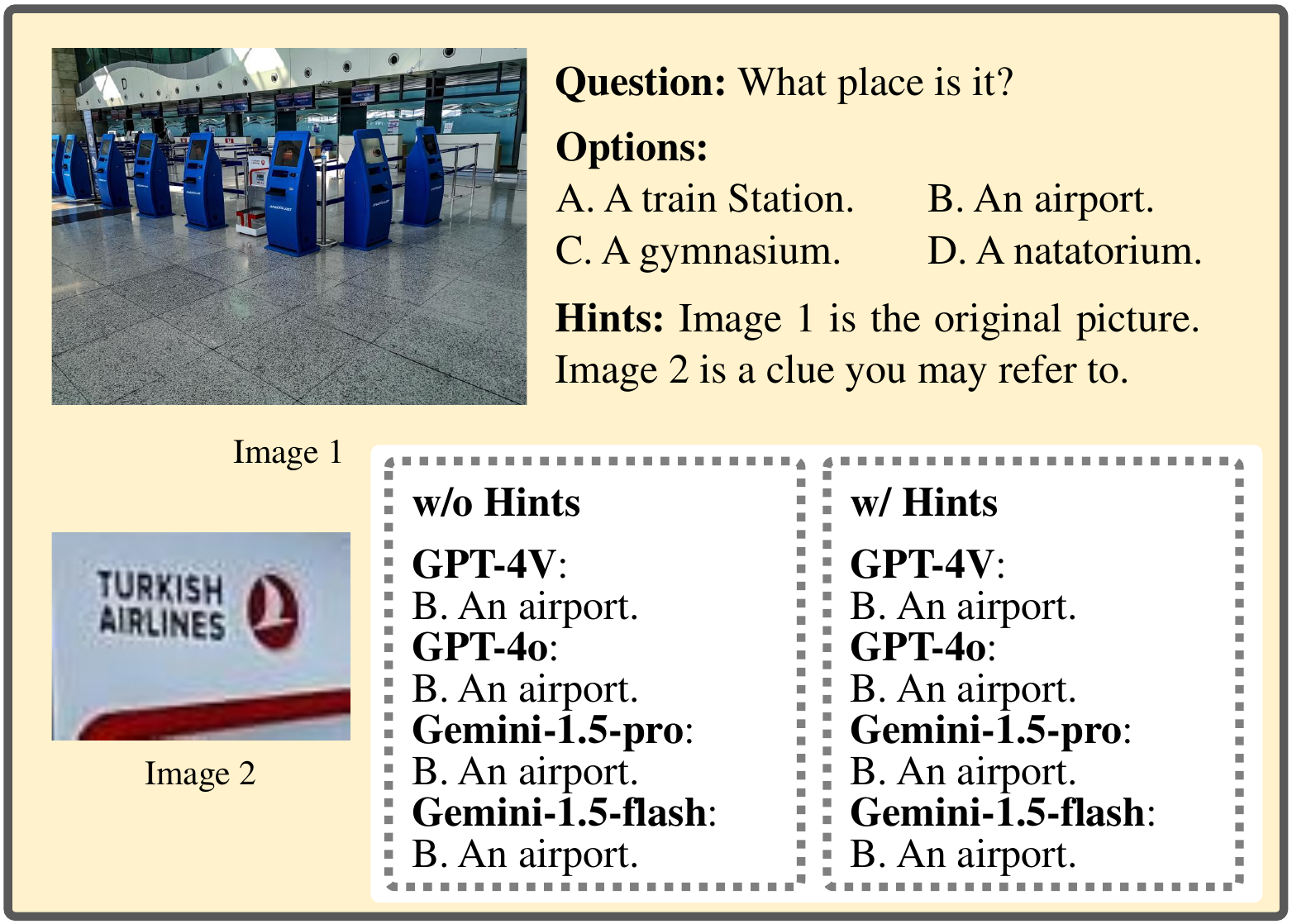}
    \end{subfigure}
    \begin{subfigure}{0.22\textwidth}
        \centering
        \includegraphics[height=4.0cm]{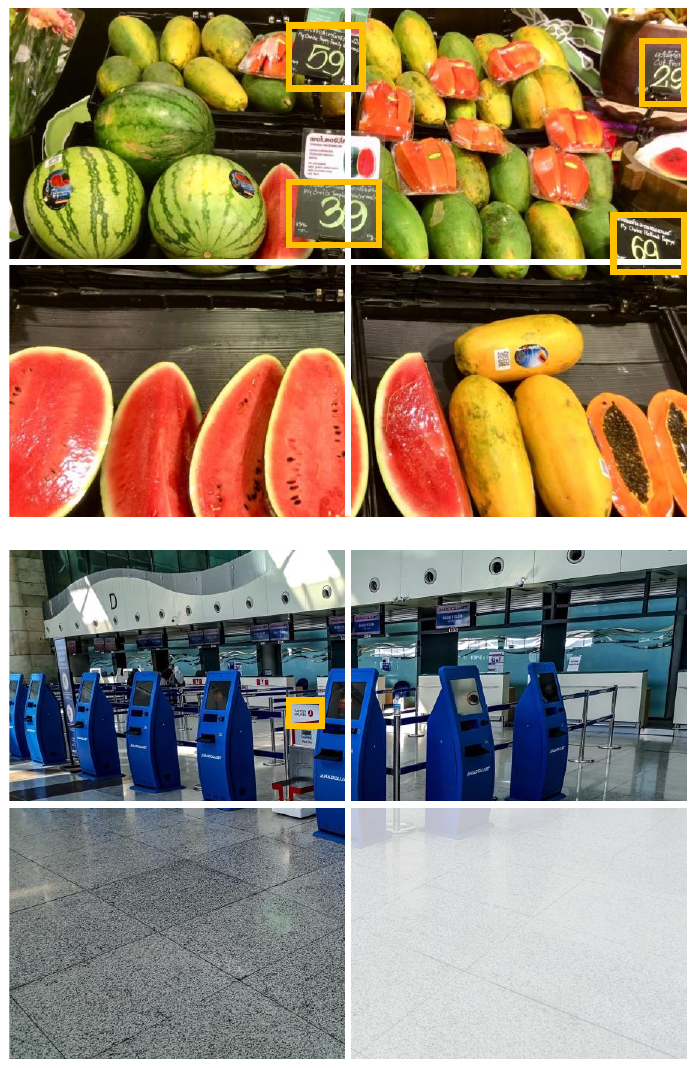}
    \end{subfigure}
    \caption{Two cases of ActiView benchmark when given human-annotated visual clues for shifting and zooming evaluation. Left: The questions and answers of models. Right: We show the location of the visual clues we provided in the original image, as well as the areas chosen by GPT-4o model. For the first case, GPT-4o chooses all the areas, and for the second case, it chooses all the areas except the one in the bottom right corner.}
    \label{fig:case-study}
\end{figure*}

In this section, we demonstrate three cases for the three proposed pipelines in Figure~\ref{fig:3cases}, and provide additional examples of integrating human-annotated visual clues hints in Figure~\ref{fig:case-study}.

\subsection{Analysis of Cases from Each Pipelines}
These results are generated by GPT-4 models and Gemini-1.5 models. Case (a) stands for the zooming evaluation, where models successfully identify the view containing useful information and generate the correct result. Case (b) illustrates a failure in the Shift-R evaluation, where all the models continue shifting to new views until all views are used. Though including the correct views, the additional views severely distract the reasoning process, where three out of four employed models produce incorrect answers.
To explore how human-like mixed evaluation affects the visual reasoning process, we further exam this failure case using GPT-4o. As shown in Figure~\ref{fig:3cases} case (c), GPT-4o first zooms into the ``upper left'' and ``upper right'' views, then discards the ``upper right'' view and shifts to the ``lower left'' one, which finally leads to the correct answer. Notably, in the final preserved views, distracting information (the highest price tag on papaya, ``69'') is screened out. This indicates that GPT-4o exhibits decent active perception abilities to move the field of view, locate details, and filter out distracting information.

\subsection{Cases of Giving Human-annotated Clues} \label{app:case}
We present a case study of ActiView in Figure~\ref{fig:case-study}. The first question targets at the most expensive watermelon, and only two out of four price tags, the ``39'' and ``59'' ones, are standing for the prices of watermelons. A distracting information appears at the ``69'' price tag that corresponds to papayas instead of watermelons. Models easily mislead by the most expensive tag ``69'' during evaluation. However, when we provide the models with the view of the price tags and remind them to focus on these tags, both GPT-4o and GPT-4V models correctly answer the question, indicating that actively perceiving key information helps improve model performance. While Gemini-1.5-pro gives the correct answer both with and without hints, and Gemini-1.5-pro fails to benefit from the hints.
The second question asks models to recognize the place of the picture. Although it may be difficult to distinguish at first glance, we can still identify this place as an airport from some details, such as a airline's logo. Since there isn't a need to extract much information from the image, and there is little distracting information, all the four models answered the question correctly both with and without hints.

The right side of Figure~\ref{fig:case-study} shows a comparison between the attention areas selected autonomously by GPT-4o and the areas highlighted by the hints we provided. It can be observed that when facing some difficult problems, although the model selects all the regions, it is unable to actively retrieve all the necessary details, thus lacking some essential information for answering the question. When the questions are relatively simple, the model successfully identify important information and gives the correct answer. This indicates that the GPT-4o model possesses a limited level of active perception capability and it still has room for improvement. We have also observed similar conclusions for other models.

\section{Prompt Template}\label{app:template}
In this section, we will provide detailed templates used for evaluation pipelines depicted in Figure~\ref{fig:pipeline}.

\subsection{Templates for General Question Answering} \label{app:template_general}
The general VQA template that requires models to answer questions given images is as following:
\begin{tcolorbox}[title=An Example Prompt for General Question Answering]
    \small 
    \texttt{Carefully analysis this image <image>, and answer the question from the given options. Question: <question> Options: <options>. Answer:} 
\end{tcolorbox}

We develop a different template for two of our evaluated models, SEAL and MGM series. These models are optimized especially on VQA tasks, and sometimes fail to strictly following long textual instructions. Therefore, we use a simple and straightforward template to prompt these models for answers as follows:
\begin{tcolorbox}[title=An Example Prompt for Question Answering(SEAL and MGM)]
    \small 
    \texttt{<question> <options>. Answer:<image>} 
\end{tcolorbox}

\subsection{Templates for Zooming Evaluation} \label{app:template_zoom}
Here are templates used in the two stages of zooming pipeline depicted in Figure~\ref{fig:pipeline} (a). Note that the term ``description\_of\_splits'' refers to the positions of the views that guide the model to shift and select views. ``description\_of\_splits'' varies depending on how the views are divided. Taking 4 sub-image for example, it is described as ``1 is the upper-left part, 2 is the lower-left part, 3 is the upper-right part, and 4 is the lower-right part.'' The model should then response with ``1, 2, 3, and/or 4'' to select the appropriate views. The prompts are as follows:
\begin{tcolorbox}[title=An Example Prompt for View Selection]
    \small 
    \texttt{This is the full image <image>, which is split in to <num\_splits> equal parts, numbered from 1 to <num\_splits>, where <description\_of\_splits>. } \\
    \texttt{===} \\
    \texttt{Response with the number of part (at least one part, at most <num\_splits> parts), that must be used to answer the question. The question is: <question>}\\
    \texttt{===} \\
    \texttt{Do not directly answer the given question. Response with the selected number of parts, split by ' if there are multiple selections. Your Response:} \\
\end{tcolorbox}

\begin{tcolorbox}[title=An Example Prompt for Zooming Question Answering]
    \small 
    \texttt{Image 0 is the full image. <zoomed\_images> These are your selected part from the full image to be zoomed for details for answering the question. Please answer question according to the given images from the the given options. Question: <question> Options: <option>. Answer:} \\ \\
\end{tcolorbox}

\subsection{Templates for Shifting Evaluation}\label{app:template_shift}
Here are templates used in the two stages of zooming pipeline depicted in Figure~\ref{fig:pipeline} (b).
\begin{tcolorbox}[title=An Example Prompt for Missing-view Examination]
    \small 
    \texttt{You will be presented with a partial image and a question concerning the full image. image 0 is <image0>, is the <image\_view> part of the full image. Given image 0, please determine if you need more visual information to answer the question: <question>} \\
    \texttt{===} \\ 
    \texttt{
Do not directly answer the question. If you can answer the question without more visual information, response with NO. Otherwise, response with other image parts you need to see given this <image\_view> part, you can choose from these views: <view\_options>. Your Response:}
\end{tcolorbox}

\begin{tcolorbox}[title=An Example Prompt for Shifting Question Answering]
    \small 
    \texttt{These are parts of an image. <all\_required\_views>. Carefully analysis these images and pay attention to their original position. Answer the question from the given options. Question: <question>. Options: <option>. Answer:} 
\end{tcolorbox}

\subsection{Templates for Mixed Evaluation} \label{app:template_mix}
Here are templates used for the mixed pipeline depicted in Figure~\ref{fig:pipeline} (c). We design two templates for regarding the type of current view. We apply template ``Operation Determination''(1) from the followings for the full images, and apply template ``Operation Determination''(2) from the followings for zoomed views. Templates are as follows:

\raggedbottom

\begin{tcolorbox}[title=An Example Prompt for Operation Determination (1)]
    \small 
    \texttt{You will be presented with a full image <image> and a corresponding question to answer. The image is split in to <num\_splits> equal parts, numbered from 1 to <num\_splits>, where <description\_of\_splits>.} \\
    \texttt{You can check for detailed visual information via zooming operation that zoom in to your selected part or parts with from the above numbers. Response with the the numbers of parts you wish to zoom in, or response with ``none'' if you don't need to can check for details. }\\
    \texttt{The quesiton is: <question> }\\
    \texttt{You should not directly answer the question. You should generate the a json dict containing 2 fields:}\\
    \texttt{- ``part'': type str, the selected numbers of index of parts, split by ``,'', or 'none' if no zooming required;}\\
    \texttt{- ``reason'': type str, why you choose these parts.} \\
    \texttt{Your response:}
\end{tcolorbox}

\begin{tcolorbox}[title=An Example Prompt for Operation Determination (2)]
    \small 
    \texttt{You will be presented with a partial image and a question concerning the full image. image 0 is <image0>, is the <image\_view> part of the full image. Given image 0, please determine if you need more visual information to answer the question: <question>} \\
    \texttt{===} \\ 
    \texttt{Your are given a full image <image> and a corresponding question to answer. The image is split in to <num\_splits> equal parts, numbered from 1 to <num\_splits>, where <description\_of\_splits>. Your have chosen to zoom in to these parts, <zoomed\_images>, for detailed checking if they can help to ansewr the quesiton. }\\
    \texttt{Question: <question> Options: <option>.}\\
    \texttt{Now, there are two operations: ``keep'' and ``shift''. }\\
    \texttt{- ``keep'': choose none or more parts from the zoomed ones to answer the question; }\\
    \texttt{- ``shift'': you can shift to the rest parts to answer questions or answer question with none sub-parts. }\\
    \texttt{You should not directly answer the question. You should return you answer in a json dict containing two fields:}\\
    \texttt{- ``zoom\_keep'': type str, the index numbers of required parts split by ``,'', or ``none'' if the zoomed parts are useless;  }\\
    \texttt{- ``shift'': type str, the index numbers of the rest parts, that are useful to the question split by ``,'', or ``none'' if you don't wish to shift.}\\
    \texttt{Your response:}
\end{tcolorbox}

\begin{tcolorbox}[title=An Example Prompt of Quesiton Ansewring for Mixed Pipeline]
    \small 
    \texttt{Image 0 is the full image. <image\_views> <image\_view\_desc> These are your selected part of image that must be used to answer the question. Please answer question according to the given images from the the given options. Question: <question> Options: <option>. Answer:} 
\end{tcolorbox}

\section{Attempts of Automatic Data Generation} \label{app:auto_gen}
In this last section, we discuss our experiments of automatic data generation, and analyse why powerful models like GPT-4V fail to accomplish this task. We will discuss the process and demonstrate typical failure cases in the following sections. 

\subsection{Automatic Data Generation Process}
In the process of automatic data generation, we used the GPT-4V model for the following experiments:
\begin{itemize}
    \item \textbf{Step 1}: We applied heuristic prompts on public datasets to encourage GPT to generate creative annotations across all types.
    \item \textbf{Step 2}: We selected the types that showed the best performance in automatic annotation and conducted batch annotation specifically for these types.
    \item \textbf{Step 3}: We manually filtered a subset of data that could be used.
\end{itemize}

In \textbf{Step 1}, we not only employed heuristic prompts to encourage GPT to generate diverse annotations but also specified the annotation types and their precise meanings (provided as candidates, encouraging the model to select from them). We restricted the annotation fields and types, and provided several manually curated examples as few-shot instances. Considering that some images in public datasets may not be suitable for our task, we allowed GPT to return ``None'' for images deemed unsuitable for annotation. The filtered annotation data were then re-evaluated using a scoring prompt, where we provided our annotation types and requirements, instructing GPT to rank the annotated data to assess its suitability.

In \textbf{Step 2}, we found that GPT performed best in annotating data of the counting type (based on a combination of manual inspection of the annotation results and GPT's automatic scoring). Therefore, we decided to use GPT for automatic annotation of counting-type data. Considering that some public datasets (such as VCR) contain images with more than one type of bounding box, we processed different bounding box types in batches for each image to ensure that only one type of object was counted at a time.

Detailed prompt templates are attached in the third sub-section of this section.

\subsection{Cases of Unsuccessful Generations of GPT-4V}
\begin{figure*}[h]
    \begin{subfigure}{0.47\textwidth}
        \centering
        \includegraphics[height=3.7cm]{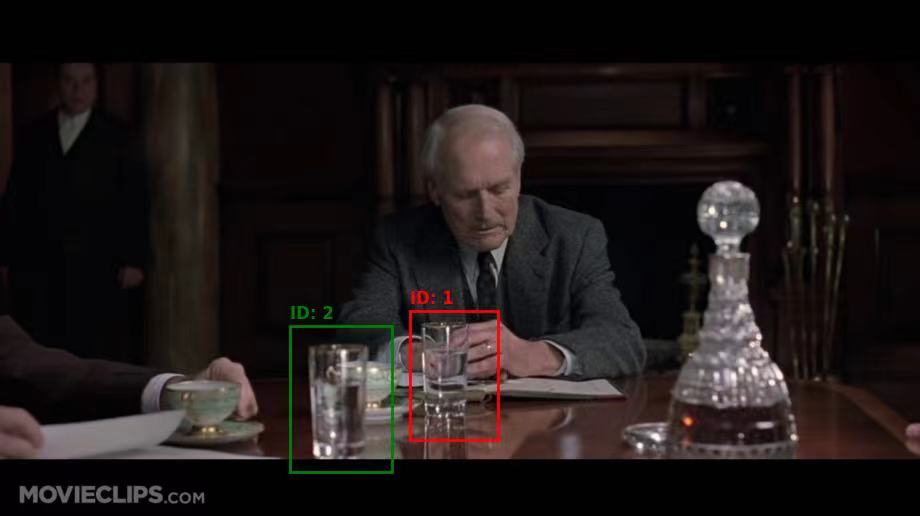}
    \end{subfigure}
    \hfill
    \begin{subfigure}{0.47\textwidth}
        \centering
        \includegraphics[height=3.7cm]{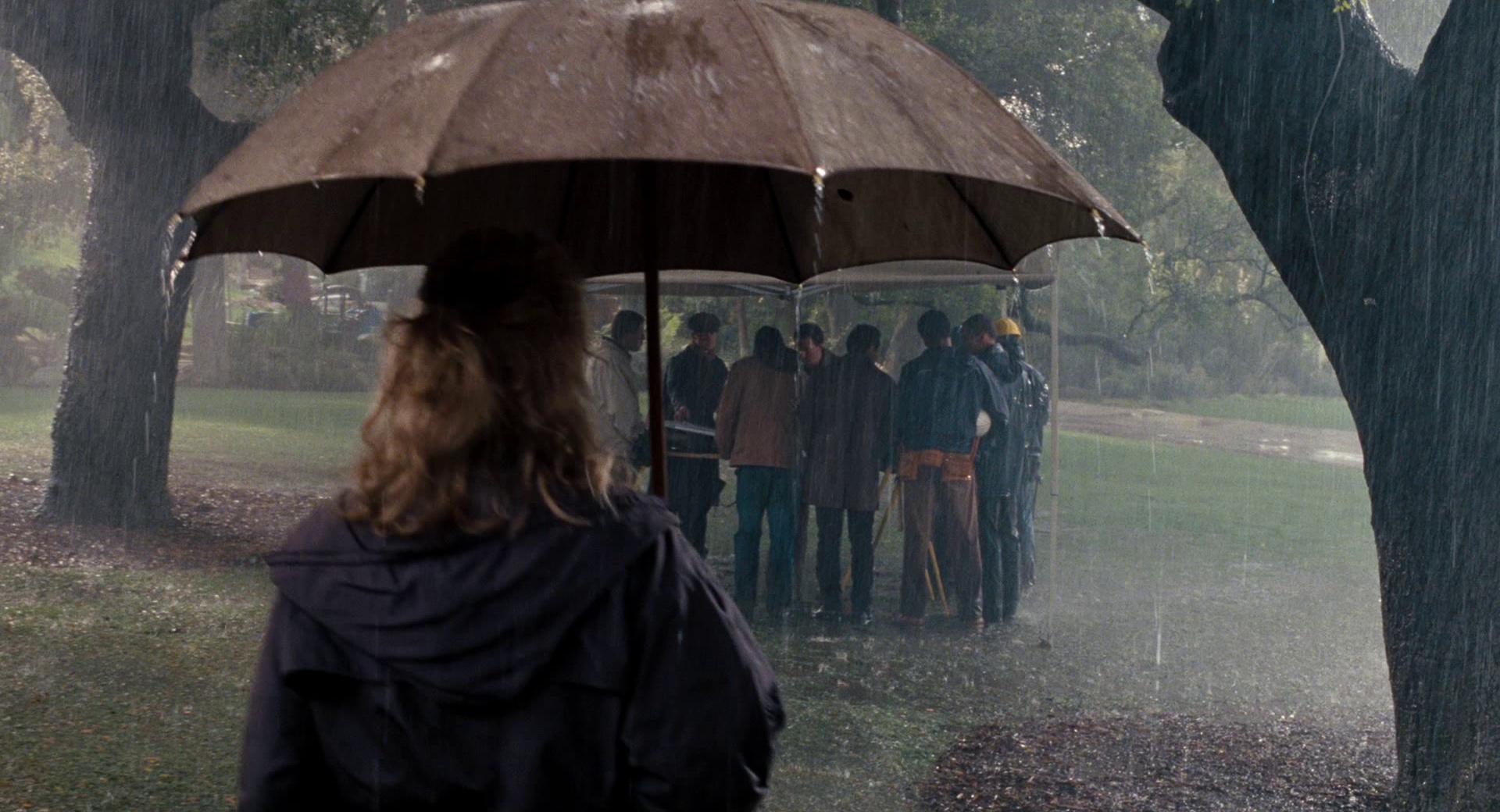}
    \end{subfigure}   
    \caption{Left: example of the automatic annotation results from Step 1. Right: example of the automatic annotation results from Step 2, where the question annotated by GPT is ``How many umbrellas can be seen in the image?''}
    \label{fig:case_gen}
\end{figure*}

We provide two typical cases demonstrating why GPT-4V fail to generate usable instances. The corresponding image is Figure~\ref{fig:case_gen}. For the case regarding the left image, it presents a typical encountered issue case of hallucination and speculation without a factual basis. Given this image, GPT-4V produces the following annotations prompted by \textbf{Step 1}:

\begin{mdframed}
    \small 
    \texttt{\{``question'': ``Which of the following best describes the setting based on the appearance and arrangement of the glass items on the table?'',} \\
    \texttt{``options'': [``A casual family dinner'', ``A quick lunch at a fast food restaurant'', ``An official or formal meeting'', ``An outdoor picnic''],}\\
    \texttt{``answer'': 2,}\\
    \texttt{``groundtruth'': ``The setting seems to be an official or formal meeting given the presence of multiple large, elegant glasses on the table, which suggest formal drinkware typically used in such settings.''\}
}
\end{mdframed}

The question and annotated answer posed by GPT-4V makes certain assumptions about the image that this scenario shows ``An official or formal meeting''. The question is not answerable concerning only this image, where it could refer to either a meeting or a dinner. Moreover, the other options except for annotated answer does not match the image in any circumstances, and can be easily eliminated without any further observation of the image. The answers does not strictly follow the given ground truth (i.e., the answer to the question cannot be rigorously inferred from the visual clues in the image), where the glasses do not support the reasoning.
For the case of the right image, it presents a typical failure case from \textbf{Step 2}. Regarding this image, GPT-4V generates an ambiguous question ``How many umbrellas can be seen in the image?'', where there are some small visible objects could potentially be umbrellas as well.

\subsection{Prompt Templates Used for Automatic Generation}
Here, we give the prompt used for automatic annotation in \textbf{Step 1} and \textbf{Step 2}.
\begin{tcolorbox}[title=Heuristic prompt used for automatic annotation in \textbf{Step 1}]
    \small 
    \texttt{The clues for marking information in several bboxes in this picture are: \{clues\}} \\
    \texttt{Based on several bboxes and corresponding clues, please design a question that requires the model to synthesize the information in these bboxes (at least two, and can only be answered based on the information in the bboxes and the clues corresponding to the annotated information). You only need to ask the question, and there is no need to repeat the clue again. } \\
    \texttt{Note that the existence of bbox (including its ID information) cannot be mentioned in the question. Questions and reasoning should be based on objective facts as much as possible instead of subjective guessing.But at the same time, you should also avoid grounding questions and questions that can be answered without pictures (including questions like what someone in the picture is doing)} \\
    \texttt{Next, mark me the corresponding information in the following format:} \\
    \texttt{1. ``question'' (str)} \\
    \texttt{2. ``options'' (list)} \\
    \texttt{3. ``abilities'' (list): choose from ``zoom in'', ``zoom out'', ``shifting'' (it is mentioned in the analysis and is not mentioned at the beginning)} \\
    \texttt{4. ``answer'' (int, index of option)} \\
    \texttt{5. ``order'': the order in which the pictures cut out of the bbox and the entire picture are displayed (the list is given in the order of reasoning, all of which are ints, representing the id corresponding to the bbox on the picture, if it is a complete picture, it is 0)} \\
\end{tcolorbox}

\begin{tcolorbox}[title=Heuristic prompt used for automatic annotation in \textbf{Step 1} (Continue)]
    \small 
    \texttt{6. “groundtruth”: Give the reasons and complete reasoning process for answering the question} \\
    \texttt{7. ``number\_of\_operations'': For example, first zoom in and then move the angle of view, it is two operations} \\
    \texttt{You must give me the answer in the following json-string format(not code block) and don\'t say anything else:} \\
    \texttt{\{\{} \\
    \texttt{  ``question'': question(str),} \\
    \texttt{  ``options'': options(list),} \\
    \texttt{  ``abilities'': ablities(list),} \\
    \texttt{  ``answer'': answer\_index(int),} \\
    \texttt{  ``order'': order(list),} \\
    \texttt{  ``groundtruth'': groundtruth(str),} \\
    \texttt{  ``number\_of\_operations'': number of the operations(int)} 
\end{tcolorbox}

\begin{tcolorbox}[title=Scoring prompt used for automatic annotation in \textbf{Step 1}]
    \small 
    \texttt{We want to design a question about the picture to test the active perception ability of the respondent. Here are the requirements:} \\
    \texttt{You will be provided with an image and information of bboxes in it. You should design a question that requires the respondent to synthesize the information in these bboxes.} \\
    \texttt{While designing the questions, you must follow these rules:} \\
    \texttt{ - The question should be based on the information in the given bboxes.} \\
    \texttt{ - The question requires the respondent to obtain information from the field of view of these bboxes as a basis, identify irrelevant information on the picture, and move the field of view of different bboxes to obtain more information before answering the question.} \\
    \texttt{ - Differences between options should be distinct. And options must not be conflict to each other.} \\
    \texttt{ - There should be one and only one correct answer among all options.} \\
    \texttt{ - The evidence or clues for answering the question must be visible in the image. } \\
    \texttt{Also, you should realize the following conditions:} \\
    \texttt{ - The answers must not require the respondent guess subjectively.} \\
    \texttt{ - You cannot generate questions require simple object grounding, e.g., what is the object in a certain region, what is the color of an object, etc.} \\
    \texttt{ - The existence of bbox and visual clues (including their ID information) cannot be mentioned in the question nor in the options.} \\
    
    \texttt{You should score the annotation through the rules given above. Here are the predefined levels for scoring, where level D is the worst and level A is the best:} \\
    \texttt{ - Level D: no reasonable questions can be generated for the given image by strictly following our rules.} \\
    \texttt{ - Level C: the question contains subjective guesses and judgments, rather than strictly following the rules(e.g. infer the location from the architectural style/image style rather than some grounding signs and texts etc.)} \\
    \texttt{ - Level B: the question can be answered via simple captioning of the pictures(like using ViT or OCR to caption the picture and ask the language model to answer the question with out the picture), or can be answered via pure common sense reasoning.} \\
    \texttt{ - Level A: the question is cleverly designed and is completely based on the information in the picture. It requires the respondents to visit different parts marked on the image for comprehensive reasoning, which fully complies with the above marking rules.} \\
\end{tcolorbox}

\begin{tcolorbox}[title=Scoring prompt used for automatic annotation in \textbf{Step 1} (Continue)]
    \small 
    \texttt{Remenber, if any subjective guess seems to appear, or anything that requires inferring from knowledge outside the image, or anything that does not follow our rule strictly (including asking for some weired questions etc.), do not hesitate to assign a low level.} \\
    
    \texttt{Here's the annotation information of the given picture:} \\
    \texttt{\{annotation\}} \\
    
    \texttt{You must give me the answer in the following json-string format(not code block) and don't say anything else:} \\
    \texttt{\{\{} \\
    \texttt{  ``score'': string, choose from ``A'', ``B'',``C'', ``D'',} \\
    \texttt{  ``reason'': string, explain why you give this score}
\end{tcolorbox}

\begin{tcolorbox}[title=Prompt used for automatic annotation in \textbf{Step 2}]
    \small 
    \texttt{You are an annotator to design questions and options for given images. Here are the guidebook for you:}
    
    \texttt{===}
    
    \texttt{Overall task description: You will be presented with an image, please generate a question, corresponding options and answer to the question, and some other information that help the reasoning process as well.}
    
    \texttt{===}
    
    \texttt{Detailed requirements you **must** follow: }
    
    \texttt{ - You must design the problem in the following type:}
    
    \texttt{   Counting with restricted information or extending reasoning based on counting. For example, there are lots of products in the image, but only a part of them are on sale, you can ask for the number of on sale products. Options are list of numbers. Candidates:}
    
    \texttt{   - How many people are wearing black hat?}
    
    \texttt{   - How many products are on discount?}
    
    \texttt{   - Which color of umbrellas are the most numerous in the picture?}
    
    \texttt{   But remember, you *cannot* ask common sense questions like how many objects are there in the picture, which can be answered without reasoning.}
    
    \texttt{ - **Simple grounding questions are NOT allowed**, such as (but not restricted to): ``what is xxx object?'', ``What is the color/style of xxx?'', and etc.}
    
    \texttt{ - For answers:}
    
    \texttt{   - By referring to the image, there must exists one and only one answer, without any ambiguities and subjective guesses. }
    
    \texttt{   - The evidence for answering the question must be visible in the image.}
    
    \texttt{   - Objective reasoning are not allowed.}
    
    \texttt{   - DO NOT rely on information that does not exist in the image.}
    
    \texttt{ - For options:}
    
    \texttt{   - The differences between generated options should be distinct. } \\
    \texttt{   - There should be one and only one correct answer among all options. }\\
    \texttt{   - Options must not be conflict to each other.} \\
    \texttt{===}\\
    \texttt{The requirements of the generated data format are as follows:}\\
    \texttt{1. ``question'' (str, start with wh words or prep + wh words)}\\
    \texttt{2. ``options'' (list)} \\
    \texttt{3. ``abilities'' (list): choose from ``zoom in'', ``zoom out'', ``shift'' } \\
    \texttt{4. ``answer'' (int, index of correction option, starting from 0)} \\
    \texttt{5. “groundtruth”: Give the reasons and complete reasoning process for answering the question} \\
    \texttt{6. ``operations'': For example, first zoom in to a region and then moving to a different region, counted as two operations} \\
    \texttt{===} \\
\end{tcolorbox}

\begin{tcolorbox}[title=Prompt used for automatic annotation in \textbf{Step 2} (Continue)]
    \small 
    \texttt{Here are some bounding boxes and their type for you to refer to:} \\
    \texttt{\{boxes\}} \\
    \texttt{The items in these bounding boxes are all \{type\}
    The questions you ask must be about the information within the bounding boxes and strictly meet the requirements and question types given to you above.} \\
    \texttt{===} \\
    \texttt{If it is impossible to come up with required questions, you should response with {{``question'': (str)``None''}} in json-string format(not code block). Otherwise, you must generate response in the following json-string format(not code block) and don\'t say anything else:} \\
    \texttt{\{\{} \\
    \texttt{  ``question'': question(str),} \\
    \texttt{  ``options'': options(list),} \\
    \texttt{  ``abilities'': ablities(list),} \\
    \texttt{  ``answer'': answer\_index(int),} \\
    \texttt{  ``order'': order(list),} \\
    \texttt{  ``groundtruth'': groundtruth(str),} \\
    \texttt{  ``operations'': number of the operations(int),} \\
    \texttt{\}\}} \\
    \texttt{===} \\
    \texttt{Please generate response for the given image that **strictly follow** the above requirments:} 
\end{tcolorbox}

\end{document}